\documentclass[11pt, letterpaper]{cmu}

\usepackage[all]{hypcap}
\usepackage[comma,numbers,sort,compress]{natbib}
\usepackage{hyperref}[citecolor=magenta]

\hypersetup{
    colorlinks = true,
    citecolor = {magenta},
}
\usepackage{url}
\usepackage{soul}

\usepackage{amsmath,amsfonts,bm}

\def\eqref#1{equation~\ref{#1}}

\def\1{\bm{1}}

\DeclareMathAlphabet{\mathsfit}{\encodingdefault}{\sfdefault}{m}{sl}
\SetMathAlphabet{\mathsfit}{bold}{\encodingdefault}{\sfdefault}{bx}{n}

\newcommand{\methodname}{{\textbf{\emph{PA-RL}}}}

\newcommand{\behavior}{{\pi_\beta}}

\newcommand{\bz}{\mathbf{z}}

\newcommand{\bs}{s}
\newcommand{\ba}{a}

\renewcommand{\mathbf}{\boldsymbol}

\newcommand{\mc}{\mathcal}

\newcommand{\bb}{\mathbb}

\makeatletter
\def\Ddots{\mathinner{\mkern1mu\raise\p@
\vbox{\kern7\p@\hbox{.}}\mkern2mu
\raise4\p@\hbox{.}\mkern2mu\raise7\p@\hbox{.}\mkern1mu}}
\makeatother

\newcommand{\brac}[1]{\left[ #1 \right]}

\numberwithin{equation}{section}

\usepackage{amsmath}
\usepackage{amssymb}
\usepackage{mathtools}
\usepackage{amsthm}
\usepackage{booktabs}
\usepackage[capitalize,noabbrev]{cleveref}
\usepackage[utf8]{inputenc} %
\usepackage[T1]{fontenc}    %
\usepackage{url} %
\usepackage{amsfonts}       %
\usepackage{nicefrac}       %

\usepackage{xspace}
\usepackage{multirow}
\usepackage{xcolor}         %
\usepackage{wrapfig}
\usepackage{bbm}
\usepackage{graphicx}
\usepackage{subcaption}
\usepackage[algoruled,boxed,lined,noend,algo2e]{algorithm2e}
\usepackage[export]{adjustbox}
\usepackage[normalem]{ulem}
\usepackage{algorithm}

\usepackage[noend]{algorithmic}

\title{Policy Agnostic RL: Offline RL and Online RL Fine-Tuning of Any Class and Backbone}

\author[1]{Max Sobol Mark}
\author[2]{Tian Gao}
\author[2]{Georgia Gabriela Sampaio}
\author[1]{Mohan Kumar Srirama}
\author[2]{Archit Sharma}
\author[2]{Chelsea Finn}
\author[1]{Aviral Kumar}

\affil[1]{Carnegie Mellon University}
\affil[2]{Stanford University}

\correspondingauthor{maxsobolmark@cmu.edu;\\ Project website (with code and additional results): \href{https://PolicyAgnosticRL.github.io/}{https://PolicyAgnosticRL.github.io/}}

\begin{abstract}
	Recent advances in learning decision-making policies can largely be attributed to training expressive policy models, largely via imitation learning. While imitation learning discards non-expert data, reinforcement learning (RL) can still learn from suboptimal data. However, instantiating RL training of a new policy class often presents a different challenge: most deep RL machinery is co-developed with assumptions on the policy class and backbone, resulting in poor performance when the policy class changes. For instance, SAC utilizes a low-variance reparameterization policy gradient for Gaussian policies, but this is unstable for diffusion policies~\citep{wang2022diffusion}
	and intractable for autoregressive categorical policies. To address this issue, we develop an offline RL and online fine-tuning approach called \textbf{policy-agnostic RL} (\methodname) that can effectively train multiple policy classes, with varying architectures and sizes. We build off the basic idea that a universal supervised learning loss can replace the policy improvement step in RL, as long as it is applied on ``optimized'' actions. To obtain these optimized actions, we first sample multiple actions from a base policy, and run global optimization (i.e., re-ranking multiple action samples using the Q-function) and local optimization (i.e., running gradient steps on an action sample) to maximize the critic on these candidates.
	\methodname{} enables fine-tuning diffusion and transformer policies with either autoregressive tokens or continuous action outputs, at different sizes, entirely via actor-critic RL. Moreover, \methodname{} improves the performance and sample-efficiency by up to \textbf{2 times} compared to existing offline RL and online fine-tuning methods. We show the first result that successfully fine-tunes OpenVLA~\citep{kim2024openvla}, a 7B generalist robot policy, autonomously with Cal-QL~\citep{nakamoto2023calql}, an online RL fine-tuning algorithm, improving from 40\% to 70\% in the real world in 40 minutes.
\end{abstract}

\begin{document}

\maketitle

\vspace{-0.25cm}
\section{Introduction}
\vspace{-0.25cm}

Recent successes in training decision-making policies in a number of domains such as robotics and language agents stem largely from the use of expressive models combined with large-scale imitation-style training~\citep{rt2,chi2023diffusion,kim2024openvla,chen2023fireact}, an approach that has been tried and tested in other areas of machine learning. However, training a policy once and freezing it is not good enough for many real-world deployment scenarios, where some adaptation is needed: for example, a robot must adapt its behavior as the surrounding environment or task changes;
a language-model powered web navigation agent must attempt to use its own experience to improve behavior as it interacts more with the world~\citep{bai2024digirl}. The hallmark of an adaptation process is in its use of autonomous, non-expert data. In these use cases, imitation alone done once or applied repeatedly is not enough to guarantee the most efficient learning. 

Reinforcement learning (RL) provides a flexible framework for adaptation and fine-tuning with non-expert data, in offline~\citep{levine2020offline}, online~\citep{nakamoto2023calql}, or hybrid~\citep{ball2023efficient} regime. In principle, off-the-shelf RL algorithms could be used to fine-tune any policy. For instance, by running actor-critic RL~\citep{suttonrlbook}, a policy can be trained towards maximizing the Q-function. However, most existing deep RL algorithms entangle the choice of training objectives and algorithm design with the choice of the policy class. For example, soft actor-critic (SAC)~\citep{sac}, the base learner for many offline and online fine-tuning algorithms~\citep{kumar2020conservative,nakamoto2023calql}, employs reprarameterization which is applicable to and stable for Gaussian (or tanh-Gaussian) policies: swapping the policy for a diffusion policy causes instability~\citep{wang2022diffusion}. These instabilities can be severe to the extent that much weaker policy extraction techniques, e.g., critic-based re-ranking~\citep{hansen2023idql,nakamoto2024steering} on top of an imitation policy can outperform the policy gradient~\citet{wang2022diffusion}, even though theoretically this is not optimal (and indeed, with Gaussian policies performs worse empirically as well~\citep{fujimoto2018off,ghasemipour2021emaq}). Likewise, in order to extend conservative Q-learning (CQL)~\citep{kumar2020conservative} to autoregressive token-based action distributions, \citet{chebotar2023q} had to make many modifications to the loss in the CQL algorithm. Overall, this means that adapting the best policy training methodologies or parameterization from one policy class to another can be challenging, and depending upon the policy itself, practitioners are forced to choose a weaker algorithm or spend cycles modifying other components of their approach.

\begin{figure}[t]
    \vspace{-0.4cm}
    \centering
    \includegraphics[width=0.93\linewidth]{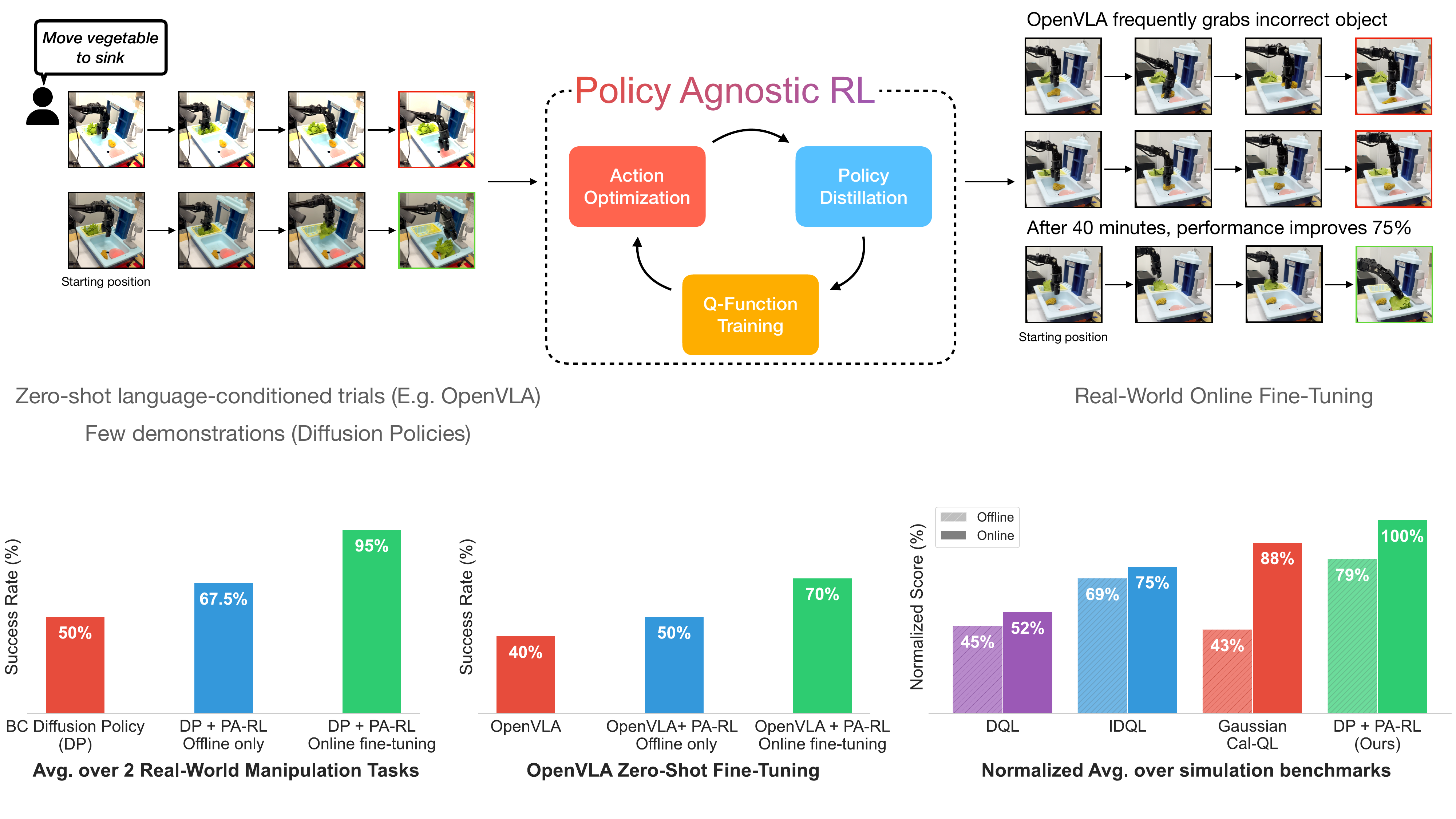}
    \vspace{-0.35cm}
    \caption{\label{fig:teaser}\footnotesize{\emph{\textbf{Policy-agnostic reinforcement learning (\methodname)}} is a simple approach for training any policy class and backbone via actor-critic RL in both the offline RL and online RL fine-tuning settings. This enables us to benefit from expressive power of different policy classes and priors from pre-training.  Our results show that \methodname{} is the first method to effectively improve diffusion policies and large generalist pre-trained policies in real-world robotic manipulation tasks. After pre-training with a few task demonstrations or zero-shot language-conditioned trials, it can significantly improve the performance of a base policy in as little as 40 minutes. On simulated benchmarks, we find substantially better results when using \methodname{} with diffusion policies, where it sets a new state-of-the-art in both offline RL and online fine-tuning, as well as autoregressive policies.}}
    \vspace{-0.25cm}
\end{figure}

We tackle this challenge by developing a single offline RL and online fine-tuning approach, which we call policy-agnostic RL (\methodname), that effectively fine-tunes any policy class or backbone. To perform policy improvement, the RL algorithm directly optimizes \emph{actions} (instead of policy parameters). Doing so decouples policy improvement from training the parameteric policy, which can now be done by maximizing the likelihood of ``optimized'' actions via supervised learning. Concretely, to obtain these optimized actions, we first sample from the base policy several times to get multiple action candidates, and then take gradient steps with respect to the value function to improve those actions in the direction of maximizing Q-values. Then these optimized action samples replace the use of samples from the policy in any value-based RL algorithm, and are used to train the policy themselves. Note that while prior work does use supervised losses for policy training, our main contribution is to show that a single approach of this sort can effectively train multiple policy classes.

We evaluate \methodname{} empirically on a number of domains including simulated robotic manipulation tasks and real robots, with Gaussian, diffusion, and autoregressive categorical policies based on transformer backbones, on offline RL and online RL fine-tuning problems. Our results show that \methodname{} achieves state-of-the-art performance, outperforming the next-best approach by 13\% in aggregate over various domains. \methodname{} produces the largest gains on long-horizon tasks that present multimodal offline data distributions (e.g., CALVIN~\citep{calvin} in our experiments), where a more expressive policy class beyond standard tanh-Gaussian is necessary for performance but has been challenging to use thus far. {Most notably, \methodname{} improves diffusion policies on two manipulation tasks by 80-100\% within only 1-2 hours of online RL fine-tuning, on a \emph{real} WidowX robot}, and improves OpenVLA~\citep{kim2024openvla}, a 7B parameter robot VLA foundation model, by 75\% after 1 hour of zero-shot trials and 40 minutes of online RL fine-tuning on the real robot.
We also provide a recommended workflow for setting knobs in \methodname{} according to the dataset and task structure, making it easy for practitioners to use our approach.

Our main contribution is \methodname{}, a \emph{single} approach for offline RL and online RL fine-tuning of policies with different classes and backbones by employing a supervised learning update on optimized actions. The use of a supervised learning loss makes our approach simple and universal. By combining global optimization and local optimization, \methodname{} is able to effectively train diffusion and transformer policies with offline RL and offline-to-online RL algorithms~\citep{nakamoto2023calql,kostrikov2021iql,ball2023efficient}. {To the best of our knowledge, our results are the first to fine-tune diffusion policies~\citep{chi2023diffusion} (both in simulation and in the real-world), and autoregressive categorical transformer policies (in simulation) and a large pre-trained robotic VLA policy, all via a single actor-critic RL approach autonomously in the real world.

\vspace{-0.3cm}
\section{Related Work}
\vspace{-0.2cm}
Contrary to prior belief, recent work~\citep{park2024value} shows that policy learning can be a big bottleneck in RL, especially in offline RL~\citep{levine2020offline}. One implication is that enhancing the policy extraction step with the most expressive architectures and the best loss functions would be important, but prior works often tailor the RL approach to a specific policy class (e.g., most work has focused on Gaussian policies). {In principle, designing effective algorithms for only one policy class can ``overfit'' resulting in methods that are actually worse for other policy classes.} For instance, while algorithms that use Gaussian policies reparameterize the policy gradient~\citep{lillicrap2015continuous,sac,fujimoto2018addressing}, doing so for diffusion policies~\citep{wang2022diffusion} or flows~\citep{mazoure2020leveraging} can be quite unstable and requires per-task tuning. {\citet{wang2022diffusion} for example requires using BC regularization and performs offline checkpoint selection against the DDPM loss. When learning with sub-optimal data, this might hurt performance.} To make a stable algorithm,
\citet{hansen2023idql} resort to Q-function re-ranking on top of a frozen behavior policy, resulting in a somewhat less powerful policy improvement operator (e.g., compared EMaQ~\citep{ghasemipour2021emaq}, which uses a similar reranking-based policy improvement operator to TD3+BC~\citep{fujimoto2021minimalist}, which optimizes the policy through the use of full policy gradient and generally performs better). Most offline RL algorithms that use autoregressive categorical transformer policies run conditional~\citep{kumar2019reward} or unconditional supervised regression~\citep{janner2021sequence,yamagata2023q,wu2024vformer}, but \citet{park2024value} show that such approaches are unable to extract the best possible policy. In fact, to fine-tune autoregressive policies directly via offline RL, \citet{chebotar2023q} had to modify value function training.

Motivated by these findings, we build a single actor-critic RL algorithm that is effective for fine-tuning arbitrary policy classes and backbones, with a focus on continuous and autoregressive token-based policies, with both diffusion and transformer backbones. Related works that fine-tune diffusion policies include: DPPO~\citep{ren2024diffusion}, which uses a two-layer diffusion-specific policy gradient loss, whereas our approach is applicable outside of diffusion policies (Section~\ref{sec:experiments}); IDQL~\citep{hansen2023idql}, which only utilizes action re-ranking akin to global optimization in \methodname{}, but does not distill it into the policy iteratively and hence results in poor fine-tuning performance in our experiments; DIPO~\citep{yang2023policy} and DDiffPG~\citep{li2024learning}, which only utilize the ``action gradient'' akin to local optimization in \methodname{}, but unlike us do so in an online setting, with no pre-training involved; and DQL~\citep{wang2022diffusion}, which utilizes the reparameterized policy gradient estimator but is quite unstable in practice, requiring specific checkpoint selection schemes and regularization to succeed, unlike our approach. \citet{psenka2023learning} learn diffusion policies via score matching, which \citet{ren2024diffusion} find to be quite unstable. Our method outperforms IDQL~\citep{hansen2023idql}, which is one of the most performant methods in this category. We also instantiate our method for fine-tuning autoregressive categorical transformer policies via offline RL and online fine-tuning methods in simulation successfully. To our knowledge, there is no prior work that attempts to fine-tune such models via value-based RL, with the exception of \citet{chebotar2023q}: unlike them, we make no modifications to value function learning.

Methodologically, our method \methodname{} appears similar to prior approaches that pose ``RL as supervised learning'', and use weighted or filtered negative log likelihood (NLL) losses for training~\citep{peng2019advantage,peters2010reps,rwr,oh2018self,mpo}. While this line of work inspires the design of our loss functions of course,  we note a crucial difference: while these works largely use the dataset or replay buffer action for training via an NLL loss, \methodname{} samples \emph{new} actions from the policy, optimizes them against the critic, and then trains the policy via NLL on this action. This allows \methodname{} to make aggressive updates, thus avoiding the ``slowness'' associated with supervised regression~\citep{tajwar2024preferencefinetuningllmsleverage,kostrikov2021iql,park2024value}, while inheriting its simplicity. In fact, \citet{tajwar2024preferencefinetuningllmsleverage} show theoretically that utilizing on-policy actions in a weighted regression loss can give rise to mode-seeking behavior akin to policy gradients, whereas using dataset actions does not exhibit this behavior. 

Action optimization from \methodname{} also resembles prior work that uses CEM optimization to obtain actions from a Q-function in the online RL setting~\citep{kalashnikov18,simmons2019q}, and supervised learning to improve a policy based on the obtained actions~\citep{neumann2023greedy,shao2022grac}. Unlike \methodname{}, these methods do not make use of offline RL pre-training to train the proposal distribution, which we find to be important since the critic can produce erroneous values outside the support of the dataset seen so far (see Figures~\ref{fig:cem-kitchen-antmaze} and \ref{fig:cem-overestimation}). Such errors can in turn hurt the efficacy of our approach.

\vspace{-0.25cm}
\section{Problem Setup and Preliminaries}
\vspace{-0.2cm}
We formalize our problem in the RL framework. The goal of RL is to find the optimal policy in an MDP, $\mathcal{M} = (\mathcal{S}, \mathcal{A}, P, r, \rho, \gamma)$, where $\mathcal{S}$ denotes the state space and $\mathcal{A}$ denotes the action space.  $P(\bs' | \bs, \ba)$ and $r(s,a)$
are the dynamics and reward functions. $\rho(s)$ denotes the initial state distribution. $\gamma \in (0,1)$ denotes the discount factor. Formally, the optimal policy in an MDP, $\pi^*:\mc S\mapsto \mc A$ maximizes the discounted sum of rewards, denoted by $V^\pi(s) = {\bb E_{\pi}\brac{ \sum_{t} \gamma^t r(s_t, a_t)|s_0=s, a_t \sim \pi(s_t), s_{t+1} \sim p(\cdot | s_t, a_t)}}$. The Q-function of a given policy $\pi$ is defined as ${Q^\pi(s,a) = {\bb E_{\pi}\brac{\sum_{t} \gamma^t r(s_t, a_t)|s_0=s,a_0=a, a_{t+1} \sim \pi(s_{t+1}), s_{t+1} \sim p(\cdot | s_t, a_t)}}}$. We use $Q_\theta^\pi$ to denote the estimate of the Q-function of a policy $\pi$ as obtained via a neural network with parameters $\theta$. The action $\ba$ is a $d$-dimensional continuous vector  in $[-1, 1]^d$.

\textbf{Problem setting.} We study two problem settings: \textbf{(a)} fully offline~\citep{levine2020offline} and \textbf{(b)} offline-to-online fine-tuning~\citep{nakamoto2023calql}. In \textbf{(a)}, we are given access to an offline dataset of experience, $\mathcal{D}_\mathrm{off} = \{(s_i, a_i, r_i, s^\prime_i)\}_{i=1}^N$, collected by a behavior policy, $\behavior$, and want to learn a policy that attains best performance using this dataset. In \textbf{(b)}, we are supposed to optimize the policy learned offline, say $\pi_\mathrm{off}$, using autonomously-collected interaction data in $\mathcal{M}$. Our goal is to obtain the optimal policy with the smallest number of online samples, efficiently. Our approach, \methodname{} prescribes a single approach to fine-tune policies of different parameterizations and classes.

\textbf{Policy classes and parameterizations.} In our experiments, we consider fine-tuning two types of policy classes: diffusion policies that produce continuous actions and autoregressive policies (based on a transformer architecture in our experiments) that produce categorical action tokens. Diffusion policies use a conditional Denoising Diffusion Probabilistic Model (DDPM,~\citet{ho2020denoising}) to represent the distribution over action conditioned on the state. A DDPM trains a diffusion step-dependant ($t$) denoising model, $\varepsilon_\phi(\ba, t | \bs)$ that is trained with:
\begin{align}
	\label{eq:diffusion_loss}
	\mathcal{L}^{\mathrm{ddpm}} (\phi) = \mathbb{E}_{t \sim \mathcal{U}(1, K), \epsilon \sim \mathcal{N}(0, I), (s, a) \sim \mathcal{D}} \left[ \| \epsilon - \epsilon_\phi ( \sqrt{\bar{\alpha}_i} a + \sqrt{1 - \bar{\alpha}_i} \epsilon, s, t ) \| \right]
\end{align}
where, given a fixed variance schedule $\beta_1, \ldots, \beta_K$ for the forward diffusion process, $\alpha_t$ is defined as $1 - \beta_t$, and $\bar{\alpha_t}$ as $\prod_{s=1}^K \alpha_s$.
To obtain the final action, we start with a random sample $\ba_K \sim \mathcal{N}(0, I)$, and iteratively denoise the sample such that $\ba_{t-1} = \frac{1}{\sqrt{\alpha_t}} \left( \ba_t - \frac{1-\alpha_t}{\sqrt{1-\bar{\alpha}_t}} \varepsilon_\phi(\ba_t, s, t)\right) + \sqrt{\beta_t} \bz$, where $\bz \sim \mathcal{N}(0, I)$ if $t>1$ and 0 otherwise, for $K$ total denoising steps. Note that while the loss in Equation~\ref{eq:diffusion_loss} is not identical to a negative log likelihood loss, it is typically derived from a lower-bound approximation to it. More importantly, we note that this loss function is the standard used for training diffusion models and is relatively well understood as opposed to using a different RL loss function for training.

In contrast, an autoregressive policy represents $\pi_\phi(\ba|\bs)$ as a product of conditional categorical distributions over each action dimension as shown below. Our experiments use a transformer architecture, along with uniform discretization into 128 bins per action dimension for simulation experiments, and the OpenVLA~\citep{kim2024openvla} tokenizer for real-world experiments, to parameterize this sort of autoregressive policy.
\begin{align}
	\pi_\phi(\ba|\bs) = \Pi_{i=1}^{d-1} \pi_\phi(\mathrm{tokenize}(\ba_i) | \bs, \ba_{0:i-1}).
\end{align}

\textbf{Offline RL and online fine-tuning methods.} The approach we build only affects policy optimization, and retains the same training procedure for the critic as the base algorithm. Our experiments will focus on two classes of actor-critic based online fine-tuning algorithms~\citep{park2024value}: \textbf{(1)} algorithms that decouple critic updates from actor updates (e.g., Implicit Q-Learning, IQL~\citep{kostrikov2021iql}), and \textbf{(2)} algorithms that sample from the actor to train the critic (e.g., Calibrated Q-Learning, Cal-QL~\citep{nakamoto2023calql}). Briefly, Cal-QL trains the Q-function to reduce temporal-difference (TD) error, with an additional regularizer that penalizes the learned Q-values on out-of-distribution (OOD) actions as long as Q-values are higher than $V^\mu(s)$, the values of a reference policy, while compensating for this pessimism on actions seen within the training dataset. The Cal-QL critic training objective is given by:
\begin{align}
	\label{eqn:cql_training}
	\mathcal{L}_Q^{\texttt{Cal-QL}}(\theta; \phi) = & \ \alpha \left(\mathbb{E}_{s \sim \mathcal{D}, a \sim \pi_\phi(\cdot|\bs)} \left[\max(Q_\theta(s,a), V^\mu(s))\right] - \mathbb{E}_{s, a \sim \mathcal{D}}\left[Q_\theta(s,a)\right]\right) \\
	                                                & + \frac{1}{2} \mathbb{E}_{s,a,s' \sim \mathcal{D}} \left[ ( Q_\theta(s, a) - \mathcal{B}^\pi \bar{Q}(s, a) )^2 \right]. \nonumber
\end{align}
Where $Q_\theta$ is the learned critic, $\bar{Q}$ is the delayed target Q-function, and $\mathcal{B}^\pi \bar{Q}(s, a)$ is the backup operator: $\mathcal{B}^\pi \bar{Q}(s, a) = r(s,a) + \gamma \bb E_{a' \sim \pi(a' | s')} [\bar{Q}(s', a')]$. Computing this loss requires sampling actions from the learned policy $\pi_\phi(\cdot|\bs)$, which is now an expressive policy class. In contrast, IQL trains the Q-function to regress to a higher expectile of the value function, without needing to query any new action samples from the learned policy (where $V_{\psi}(s)$ is the value network).
\begin{align}
	\label{eq:iql_equation}
	\mathcal{L}_V^\mathrm{IQL} (\psi) = \mathbb{E}_{(s, a) \sim \mathcal{D}} \left[L_2^\tau (Q_{\hat{\theta}}(s, a) - V_\psi(s))\right] \\
	\mathcal{L}_Q^\mathrm{IQL}(\theta) = \mathbb{E}_{(s, a, s') \sim \mathcal{D}} \left[ (r(s, a) + \gamma V_\psi(s') - Q_\theta(s, a))^2 \right]
\end{align}
Where $L_2^\tau(u) = |\tau - \mathbbm{1}(u < 0)| u^2$ is the expectile loss, and $\hat{\theta}$ are the target parameters for the Q-function. Prior algorithms for  diffusion policies largely do not apply to autoregressive policies as they make design choices specific to the diffusion process: for example, \citet{ren2024diffusion} exploits the structure of diffusion.

\vspace{-0.2cm}
\section{Policy Agnostic RL (\methodname): Training Multiple Policy Classes with Actor-Critic RL}
\label{sec:method}
\vspace{-0.2cm}

Our approach aims to fine-tune multiple policy classes with RL, regardless of scale, class and output type, stably and efficiently. A prevalent approach to attain sample-efficient policy improvement is to use an off-policy RL method, which typically alters between fitting an action-value  Q-function and updating the policy parameters in the direction of larger predicted Q-values. Typically, value learning treats the policy as a black-box that provides actions for computing and optimizing the Bellman update. Policy improvement, on the other hand, requires optimizing the value function with respect to the policy parameters. For example, most continuous action RL algorithms estimate the gradient $\nabla_\phi Q(s, \pi_\phi (s))$ with respect to the parameters of the policy $\phi$ for this purpose. Unfortunately, estimating this gradient can be tricky for several policy classes. {For e.g., for large diffusion policies propagating the policy gradient through the denoising chain can be unstable, often requiring extensive per-environment tuning of hyperparameters~\citep{wang2022diffusion} or truncating the gradient propagation after a subset of denoising steps~\citep{ren2024diffusion}.}
Similarly, for auto-regressive policies that operate on discrete action tokens, we must utilize a high-variance REINFORCE~\citep{williams1992simple} policy gradient to optimize the policy. This is not desirable.

\textbf{Can we devise a simple and practically feasible, yet universal approach to policy optimization in offline RL and online fine-tuning?} One approach is to use a loss function that is universally applicable to most deep learning machinery, such as the supervised learning loss: e.g., a negative log likelihood (NLL) loss or its approximation, such as the variational lower bound~\citep{ho2020denoising}. \textbf{\emph{Our method (Fig.~\ref{fig:method}) builds on the idea}} that policy improvement can be performed via such a loss, as long as the loss is applied on \emph{optimized} actions. Hence, we can decompose the policy improvement step in two stages: \textbf{(1)} directly optimizing action samples produced by the policy, and \textbf{(2)} training the policy to imitate these ``optimized'' actions. This decomposition avoids needing to compute $\nabla_\phi Q(s, \pi_\phi (s))$, or estimating high-variance policy gradient estimates. Since policy improvement is decoupled from policy training, we refer to this approach as \textbf{\emph{``policy-agnostic RL''}} or \methodname{} in short. We would expect this approach to inherit appealing attributes pertaining to scaling, reliability, and easy tuning of supervised learning losses. In this section, we will detail each of the two stages of our approach, and then describe the final algorithm.

\begin{figure}[t]
    \vspace{-0.4cm}
    \centering
    \includegraphics[width=0.75\linewidth]{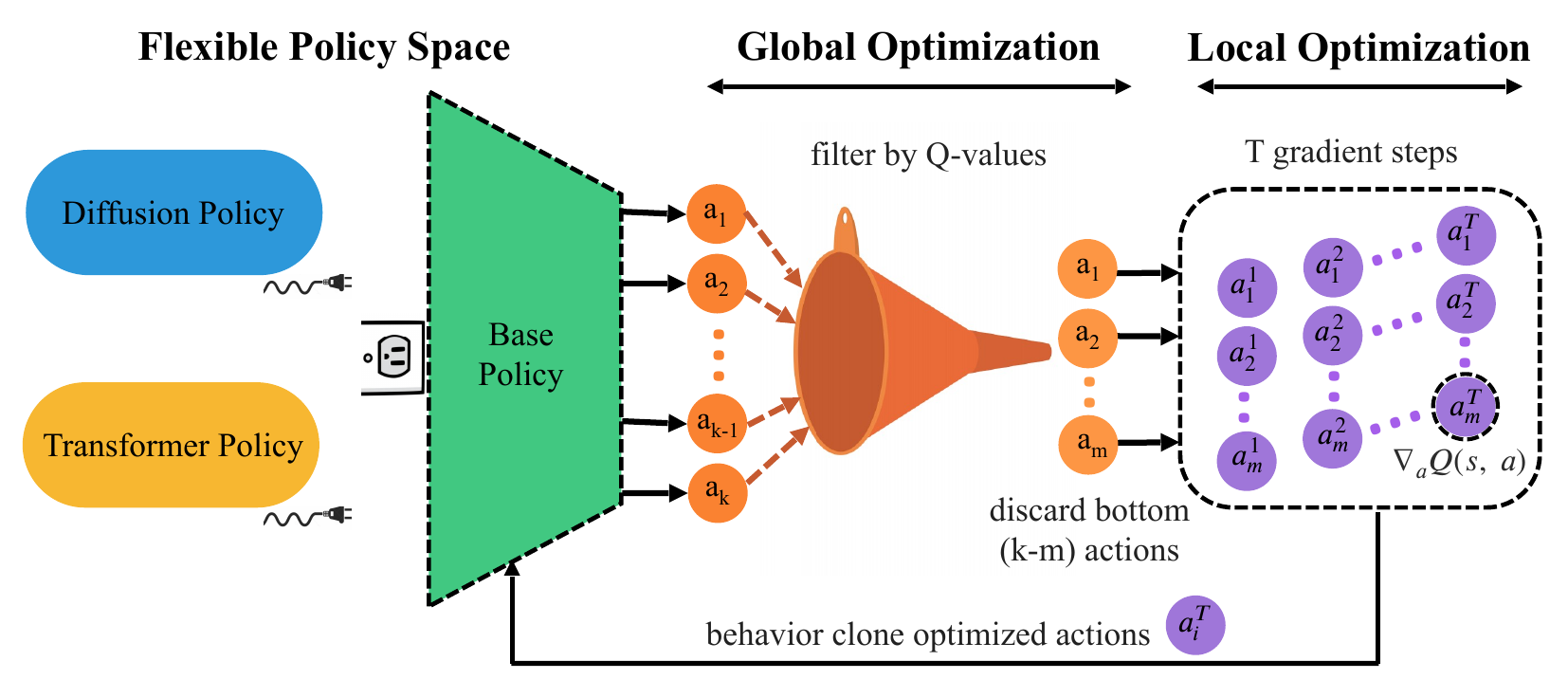}
    \vspace{-0.2cm}
    \caption{\label{fig:method}\footnotesize{\emph{\textbf{An overview of \methodname{}.}} Instead of directly passing critic gradients through the policy parameters, \methodname{} first ``optimizes'' actions via critic re-ranking and gradient ascent. Then, it trains the policy to mimic the most optimized action.}}
    \vspace{-0.25cm}
\end{figure}

\vspace{-0.2cm}
\subsection{Stage I: Action Optimization}
\vspace{-0.2cm}
Given a state $s$, a policy $\pi_\phi (\cdot|s)$, and a fixed Q-function $Q_\theta (s, a)$, the objective of this stage is to obtain an action sample that optimizes the Q-function while not deviating too far from the support of seen actions at state $s$. We use $\pi_\phi(\cdot | s)$ as an initializer for the action optimization procedure. In the offline setting, doing so allows us to find the best action within the support at the current state, when the critic is conservative~\citep{kumar2020conservative,nakamoto2023calql}. During fine-tuning, this enables us to still leverage priors learned by the offline policy while adapting it to maximize returns on the task.

To produce an optimized action, we utilize a combination of different types of \emph{action optimization} procedures. First, we consider \emph{\textbf{global}} optimization that samples multiple actions from the pre-trained policy, followed by discarding all but the top few actions with the highest Q-values under the trained critic for computational efficiency. Formally, let $\mathcal{A}_{\pi_\phi, k}(\bs): = \{\ba_0, \ba_1, \cdots, \ba_{k-1} \} \sim \pi_\phi(\cdot|\bs)$ denote $k$ sampled actions from the policy. And let $\widetilde{\mathcal{A}}_{\pi_\phi, k}(\bs) := \{\ba[0], \ba[1], \cdots, \ba[k-1]\}$ denote the set $\mathcal{A}_{\pi_\phi, k}(\bs)$ with actions put in order of their ranking obtained from the Q-function, i.e., $Q_\theta(\bs, \ba[i]) \geq Q_\theta(\bs, \ba[j])$, for $i \leq j$. Then, global optimization retains the following subset:
\begin{align}
    \label{eq:global optimization}
    \widetilde{\mathcal{A}}_{\pi_\phi, m}(\bs) = \{\ba[0], \ba[1], \cdots, \ba[m-1] \}, ~~m \leq k. ~~~~~~~ \textbf{(global optimization)}
\end{align}
Given this subset of the top $m$ actions at a state $\bs$, we now \emph{\textbf{locally}} improve each action, by performing gradient steps on the action in the direction of the gradient of the Q-function, directly on the action itself, without changing the policy parameters at all. This sort of a fine-grained local optimization is complementary to the fairly coarse global optimization procedure above as it perturbs the action to another one in its vicinity. Formally, given an action sample $\ba[i]$, we run $T$ steps of gradient ascent starting from $\ba^0[i] := \ba[i]$ to obtain the locally optimal action, $\ba^T[i]$ as shown below.
\begin{align}
    \label{eq:local_optimization}
    \text{for}~ j=0, \cdots, T-1,~~ \ba^{j+1}[i] = \ba^j[i] + \alpha \nabla_{\ba} Q_\theta(\bs, \ba) \big\vert_{\ba = \ba^j[i]}, ~~~~~~~ \textbf{(local optimization)},
\end{align}
where $\alpha$ is an appropriate learning rate that we choose for optimization. Applying both of these steps enables action optimization to leverage complementary benefits of both of these steps, while avoiding failure modes of either approach (e.g., being trapped in local minima vs not being fine-grained enough). Concretely, let us denote the action set obtained by running local optimization on $\widetilde{\mathcal{A}}_{\pi_\phi, m}(\bs)$ as $\widetilde{\mathcal{A}}^T_{\pi_\phi, m}(\bs)$. A pseudocode for action optimization is in Algorithm~\ref{alg:act_optimization}.

\vspace{-0.2cm}
\subsection{Stage II: Policy Training via Supervised Learning}
\vspace{-0.2cm}
The second stage of \methodname{} distills optimized actions into the learned policy model. Crucially, this distillation is performed via standard likelihood maximization procedures from supervised learning that most deep learning models are trained to do (or optimization of the standard lower-bound on likelihood for diffusion models). While the most direct option is to simply take the action from the set $\widetilde{\mathcal{A}}^T_{\pi, m}(\bs)$ that attains the highest Q-value (say, $\ba^*(\pi, m, T, \bs)$, where the arguments correspond to various design knobs of action optimization) and maximize its likelihood under the learned policy $\pi_\phi(\cdot|\bs)$, another alternative is to distill all action samples from $\widetilde{\mathcal{A}}^T_{\pi_\phi, m}(\bs)$, but weight the contributions of different actions using the Q-value. We prescribe a simple strategy to choose between these methods (Appendix~\ref{appendix:parl-details}).
To accomplish this, we define a categorical policy distribution over the optimized action samples:
\begin{align}
    \label{eq:optimized_policy}
    \pi_\phi^\mathrm{Opt}(\ba|\bs, m) := \mathbb{I} \left[ \ba \in \widetilde{\mathcal{A}}^T_{\pi_\phi, m}(\bs) \right] \cdot \frac{\exp (Q_\theta(\bs, \ba))}{\sum_{\ba' \in \widetilde{\mathcal{A}}^T_{\pi_\phi, m}(\bs)} \exp (Q_\theta(\bs, \ba'))},
\end{align}
and train the policy $\pi_\phi(\cdot|\cdot)$ to match this distribution. To do so, we annotate all states in the dataset (including the replay buffer in online fine-tuning) with an action sample from $\pi_\phi^\mathrm{Opt}(\ba|\bs, m)$, and maximize the likelihood of these actions under the policy, following best practices for supervised learning on this policy class. Formally, we denote this dataset of optimized actions as: 
\begin{align}
    \mathcal{D}^\mathrm{Opt}_{(\phi, \theta, m)} = \left\{\left(\bs_i, \tilde{\ba}^\mathrm{Opt}_i\right), ~~ \tilde{\ba}^\mathrm{Opt}_i \sim \pi_\phi^\mathrm{Opt}(\ba|\bs_i, m) \right\}_{i=1}^N.
\end{align}
For instance, if the policy $\pi_\phi$ is parameterized as a diffusion model, we follow the DDPM~\citep{ho2020denoising} behavior cloning (BC) objective, and train the policy to predict noise:
\begin{align}
\label{eqn:ddpm_loss_iql}
\mathcal{L}_\mathrm{policy}^{\mathrm{ddpm}} (\phi; \theta) = \mathbb{E}_{t \sim \mathcal{U}(1, T), \epsilon \sim \mathcal{N}(0, I), \textcolor{red}{(s, a) \sim \mathcal{D}^\mathrm{Opt}_{(\phi, \theta, m)}}} \left[ \| \epsilon - \epsilon_\phi ( \sqrt{\bar{\alpha}_i} a + \sqrt{1 - \bar{\alpha}_i} \epsilon, s, t ) \| \right]
\end{align}
By using this loss instead of the reparameterized Q-function gradient, we avoid ever backpropagating through the denoising chain, and instead supervise every step of the chain independently. For auto-regressive transformer policies, we use cross-entropy loss objective for next-token prediction.

Finally, we would like to note that while prior work does explore supervised learning losses for training policies~\citep{peng2019advantage,mpo,oh2018self}, the crucial differences between \methodname{} and these prior techniques stem from the fact that action samples are drawn from the \emph{current} policy, instead of a previous policy or a behavioral policy~\citep{peng2019advantage}, which enables actions to deviate farther from the data manifold. That said, since these action particles are still drawn from the \textit{current} snapshot of the learned policy, we are also able to ensure that global and local optimization do not move the actions too far away from the data manifold which can be problematic in offline RL settings. We show in our experiments that they have a substantial impact on performance and efficiency of RL training. Concretely, we outperform methods that use advantage-weighted regression (AWR) for policy extraction since \methodname{} deviates farther away from the data manifold as well as CEM-based policy extraction~\citep{kalashnikov18,simmons2019q} which falls prey to out-of-distribution actions since it finds action particles that maximize the critic in any region of the action space.

\noindent
\begin{minipage}{0.49\linewidth} %
    \vspace{-0.2cm}
    \begin{algorithm}[H]
    \caption{\label{alg:act_optimization}Action Optimization $\pi_{(\phi, \theta)}^{\mathrm{opt}}$}
    \begin{algorithmic}[1]
    \REQUIRE base policy $\pi_\phi$, Q-function $Q_\theta$
    \STATE Sample actions from $\pi$ to obtain $\mathcal{A}_{\pi_\phi, k}(\bs)$.
    \STATE Run global optimization for every state $\bs$ to retain top $m$ actions, $\widetilde{\mathcal{A}}_{\pi_\phi, m}(\bs)$
    \FOR{$a$ in $\widetilde{\mathcal{A}}_{\pi_\phi, m}(\bs)\cup \{a_\mathrm{data}(\bs)\}$}
    \FOR{i in \{1, \ldots, T\}}
    \STATE $a^{(i)} \gets a^{(i-1)} + \alpha \nabla_{a} Q_\theta (s, a^{(i-1)})$
    \IF{$Q_\theta(s, a^{(i)}) \leq Q_\theta(s, a^{(i-1)})$}
    \STATE $a^{(i)} \gets a^{(i-1)}$
    \ELSE
    \STATE \textbf{Break}
    \ENDIF
    \ENDFOR
    \ENDFOR
    \STATE \Return $\pi_{(\phi, \theta)}^{\mathrm{opt}}$ computed via Equation~\ref{eq:optimized_policy}
    \end{algorithmic}
    \end{algorithm}
    \end{minipage}
~\vline~
\begin{minipage}{0.49\textwidth}
    \vspace{-0.2cm}
\begin{algorithm}[H]
\caption{Cal-QL + \methodname}
\label{alg:calql_plus_ours}
\begin{algorithmic}[1]
\REQUIRE BC loss $\mathcal{L}_\textrm{policy}$, e.g. $\mathcal{L}_\textrm{policy}^\textrm{ddpm}$
\STATE Pre-train policy $\pi_\phi$ via offline RL / BC
\STATE Initialize Q-function $Q_\theta$
\FOR{step $t$ in \{1, \ldots, M\}}
\STATE Train Q-function using Eq. \ref{eqn:cql_training}, but use optimized actions for TD targets
\begin{equation}
    \theta_t = \theta_{t-1} - \eta_Q \nabla_\theta \mathcal{L}_Q^{\textrm{Cal-QL}}(\theta; \phi) \nonumber
\end{equation}
\STATE Distill optimized actions to policy
\begin{equation}
    \phi_t = \phi_{t-1} + \eta_\pi \nabla_\phi \mathcal{L}_{\textrm{policy}} (\phi; \theta)\nonumber
\end{equation}
\STATE \textbf{Collect new online rollouts:}
\STATE ~~~~$a_t \sim \pi_{(\phi, \theta)}^\textrm{opt}; s_{t+1} \sim p(s_{t+1} | s_t, a_t)$
\STATE ~~~~$\mathcal{D} \gets \mathcal{D} \cup \{(s_t, a_t, r(s_t, a_t), s_{t+1})\}$
\ENDFOR
\end{algorithmic}
\end{algorithm}
\end{minipage}

\vspace{-0.25cm}
\subsection{Putting it All Together: Final \methodname{} Algorithm}
\vspace{-0.2cm}
\methodname{} can be used to replace the policy improvement step in multiple RL algorithms. We primarily focus on online fine-tuning and adaptation of offline RL. Hence, we instantiate \methodname{} using two popular RL fine-tuning methods: Cal-QL~\citep{nakamoto2023calql} and IQL~\citep{kostrikov2021iql}. \methodname{} only modifies the policy improvement step of each of these methods, while keeping the critic training as is. Since IQL training does not utilize policy backups, using \methodname{} in conjunction with IQL is straightforward: simply replace the advantage-weighted regression (AWR) update with the above supervised learning update (e.g., Equation~\ref{eqn:ddpm_loss_iql} for diffusion policies). On the other hand, for Cal-QL and other actor-critic algorithms, where the policy $\pi_\phi(\cdot|\bs)$ is used to generate action samples for performing the TD-backup, we utilize the optimized action set $\widetilde{\mathcal{A}}^T_{\pi_\phi, m}$ for the Bellman backup. Formally, this means that instead of computing Bellman targets using an updated $\pi_\phi$, we simply compute targets using the optimized policy $\pi^\mathrm{Opt}_\phi(\cdot|\cdot, m)$ (Equation~\ref{eq:optimized_policy}) for Cal-QL. A pseudocode of the algorithm along with the corresponding changes in red is shown in Algorithm \ref{alg:calql_plus_ours}.

\textbf{Implementation details.} We provide a detailed list of hyperparameters and best practices for running \methodname{} in Appendix~\ref{appendix:parl-details}. We run \methodname{} with both state-based and image-based environments, where we utilize best design practices for the critic~\citep{kumar2022pre}. We also find that additionally including the action $a$ appearing at a given state in the dataset into action optimization can sometimes be helpful. Finally, since native gradient ascent for local optimization is not guaranteed to improve the Q-value for a larger than ideal step size, we only execute a local update if it increases the Q-value after that step.

\vspace{-0.2cm}
\section{Experimental Evaluation}
\label{sec:experiments}
\vspace{-0.2cm}
The goal of our experiments is to understand the efficacy of \methodname{} in fine-tuning policies of various parameterizations, classes, and types via RL. To this end, we evaluate \methodname{} and several prior approaches, in a number of benchmark domains that require learning policies from static offline data (offline RL~\citep{levine2020offline}) and then fine-tune them with limited online interaction in the MDP (offline-to-online fine-tuning~\citep{nair2020accelerating}). We also study the hybrid RL problem setting (i.e., online RL with offline data put in the replay buffer)~\citep{song2023hybrid,ball2023efficient} for some experiments. Then, we will also present results validating the efficacy of \methodname{} on three real-robot manipulation tasks. Finally, we perform ablation experiments to understand the utility of different components of \methodname. We first describe our main results and then present ablations.

\vspace{-0.2cm}
\subsection{Results: Simulated Benchmarks from State and Image Observations}
\vspace{-0.2cm}
We first compare \methodname{} with prior methods in the D4RL~\citep{fu2020d4rl} suite. Since we report performance in both the offline RL and offline-to-online RL settings, we apply \methodname{} on top of Cal-QL~\citep{nakamoto2023calql} and IQL~\citep{kostrikov2021iql}, two common offline RL and offline-to-online fine-tuning algorithms, although most of our results use Cal-QL. We first demonstrate the efficacy of \methodname{} in training diffusion policies and compare it to methods that train diffusion policies. Specifically, we compare \methodname{} to: \textbf{(1)} Implicit Diffusion Q-Learning (IDQL,~\citet{hansen2023idql}), which extends IQL to use diffusion policies via critic-based reranking; \textbf{(2)} Diffusion Policy Policy Optimization (DPPO,~\citet{ren2024diffusion}), which fine-tunes diffusion policies learned via imitation learning using PPO~\citep{schulman2017proximal}; and \textbf{(3)} Diffusion Q-Learning (DQL,~\citet{wang2022diffusion}), which trains diffusion policies via a reparameterized policy gradient estimator akin to standard SAC~\citep{haarnoja2018soft}. 

\textbf{Domains and tasks.} We study: \textbf{(1)} $\mathrm{AntMaze}$ tasks from D4RL~\citep{fu2020d4rl} that require controlling the joints of a quadruped ant to reach a goal location in four different maze layouts. A sparse binary reward is given upon reaching the goal; \textbf{(2)} $\mathrm{FrankaKitchen}$ tasks~\citep{gupta2019relay}, which require solving a sequence of four manipulation tasks in a kitchen environment with a 9-Dof Franka robot; and \textbf{(3)} the $\mathrm{CALVIN}$ benchmark~\citep{calvin,shi2022skill} (D $\rightarrow$ D, with distractor objects), which requires solving a sequence of four manipulation tasks in a tabletop environment. All of these tasks present long horizons; the FrankaKitchen and $\mathrm{CALVIN}$ tasks require chaining different skills into a coherent long episode. That said, the $\mathrm{CALVIN}$ task is substantially harder than FrankaKitchen since policies must be learned directly from pixels and with offline play data generated via human teleoperation. This offline data presents fairly low action coverage but pretty high coverage over different modes of semantic behavior. Due to the diversity of offline data, we believe that $\mathrm{CALVIN}$ should stress test the ability of any approach to effectively utilize the multi-modal nature of diffusion policies for improving efficiency of fine-tuning. More details about these tasks are in Appendix~\ref{appendix:env-details}.

\emph{\textbf{Results: \methodname{} significantly improves learning efficiency and asymptotic performance of Cal-QL with diffusion policies.}} We compare different approaches for offline RL training and online fine-tuning in  Table~\ref{tab:offline-online-benchmarks} and present corresponding learning curves in Figure~\ref{fig:paper_plots}. First, observe that \methodname{} attains higher offline performance than other methods that use diffusion policies, as well as standard Cal-QL with a tanh-Gaussian policy. Fine-tuning from the offline RL policy learned by \methodname{} also leads to the best fine-tuned performance in aggregate across all methods. Concretely, the fine-tuning performance of \methodname{} is 13\% higher than the next best method. In the hardest task $\mathrm{CALVIN}$ (where we must learn to control policies from raw visual observations), \methodname{} attains a \textbf{69\% improvement} over the \emph{next best} method. This perhaps hints at the efficacy of \methodname{} in effectively leveraging the increased capacity and expressive power of diffusion policies. 
Diving deeper, the learning curves in Figure~\ref{fig:paper_plots} reveal a much stronger trend: the performance of \methodname{} largely stays above the performance of all other methods throughout training.
We also evaluate \methodname{} in conjunction with IQL on the FrankaKitchen tasks in Table~\ref{tab:rlpd-kitchen}, and observe that \methodname{} + IQL also outperforms standard IQL. Hence, \methodname{} is broadly effective.

\begin{figure}[t]
\vspace{-0.2cm}
    \centering
    \includegraphics[width=0.97\textwidth]{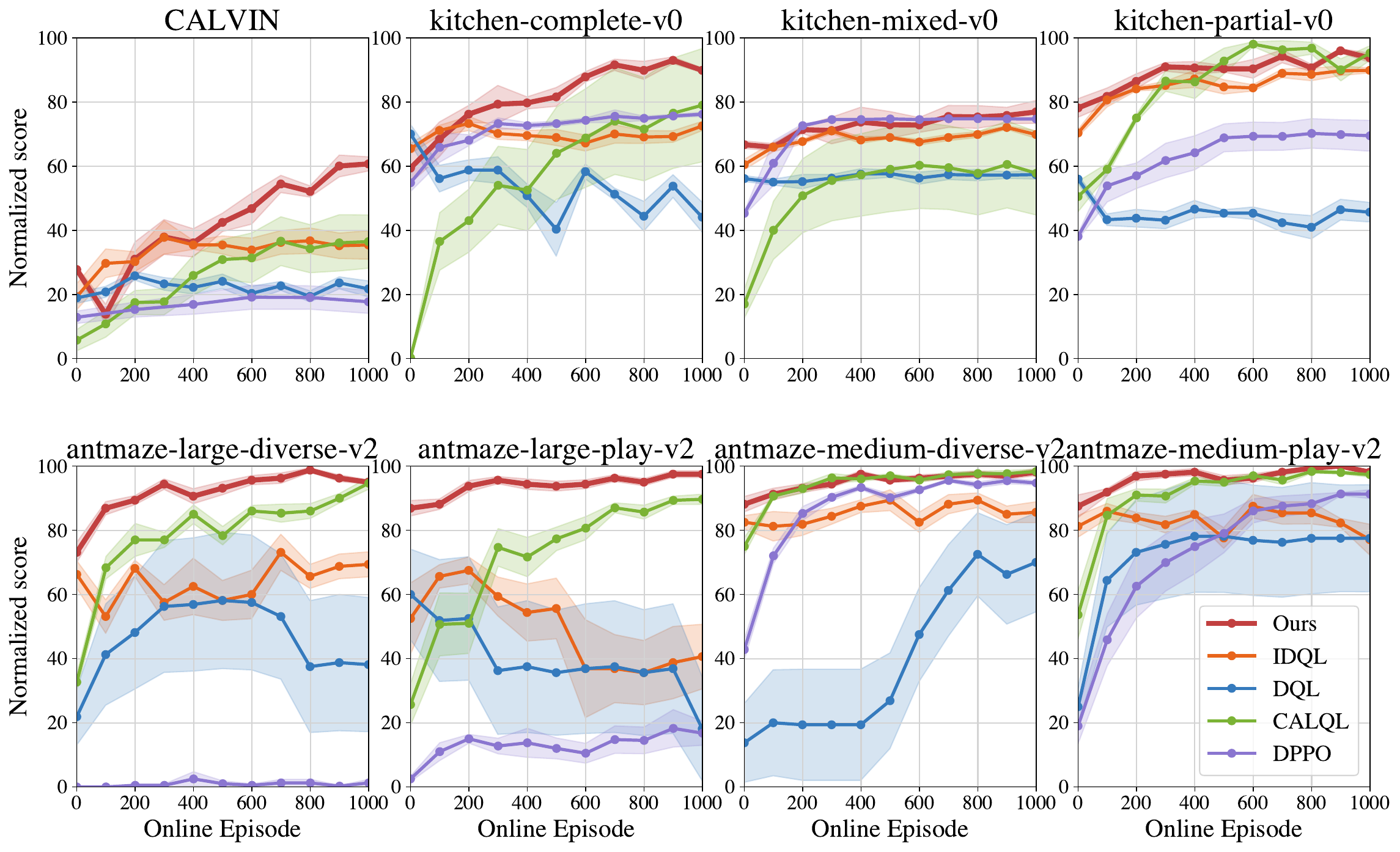}
    \vspace{-0.4cm}
\caption{\footnotesize{\textbf{Learning curves of online fine-tuning} with various methods. Observe that \methodname{} + Cal-QL (red) largely always dominates or attains similar performance to the next best method. Other methods for fine-tuning diffusion policies (IDQL, DQL, DPPO) are a bit unstable, and perform substantially worse.} Since DPPO is substantially more data inefficient, we plot it with different x-axis units: for kitchen each unit is 500 episodes (axis goes from 0 to 500k), for antmaze each unit is 100 episodes (axis goes from 0 to 100k) and for calvin each unit is 10 episodes (axis goes until 10k). }
\label{fig:paper_plots}
\vspace{-0.2cm}
\end{figure}

\begin{table}[t]
\centering
\resizebox{0.90\linewidth}{!}{\begin{tabular}{lcccc|c}
\toprule
\textbf{Domain / Task} & \textbf{IDQL} & \textbf{DQL} & \textbf{DPPO} & \textbf{Cal-QL} & \textbf{\methodname{} + Cal-QL (Ours)} \\
\midrule
\textit{\textbf{CALVIN}} & 19 $\to$ 35& 19 $\to$ 22& 13 $\to$ 18 & 6 $\to$ 36 & 28 $\to$ 61\\
\midrule
\multicolumn{6}{l}{\textit{\textbf{Kitchen} (-v0)}} \\
\hspace{2mm}complete&  65 $\to$ 72& 70 $\to$ 44& 55 $\to$ 76& 19 $\to$ 57& 59 $\to$ 90\\
\hspace{2mm}mixed& 60 $\to$ 70& 56 $\to$ 57& 45 $\to$ 75&  37 $\to$ 72& 67 $\to$ 77\\
\hspace{2mm}partial& 70 $\to$ 90& 56 $\to$ 46& 38 $\to$ 69&  59 $\to$ 84& 78 $\to$ 94\\
\midrule
\multicolumn{6}{l}{\textit{\textbf{Antmaze} (-v2)}} \\
\hspace{2mm}large-diverse& 66 $\to$ 69& 22 $\to$ 38& 0 $\to$ 1& 33 $\to$ 95& 73 $\to$ 95\\
\hspace{2mm}large-play& 53 $\to$ 41& 60 $\to$ 18& 2 $\to$ 17& 26 $\to$ 90& 87 $\to$ 98\\
\hspace{2mm}medium-diverse& 83 $\to$ 86& 14 $\to$ 70& 43 $\to$ 95& 75 $\to$ 98& 88 $\to$ 98\\
\hspace{2mm}medium-play& 81 $\to$ 77& 25 $\to$ 78& 19 $\to$ 91& 54 $\to$ 97& 88 $\to$ 98\\
\midrule
\textbf{Aggregate} & 497 $\to$ 540 & 322 $\to$ 373 & 215 $\to$ 442& 309 $\to$ 629 & 568 $\to$ 711\\
\bottomrule
\end{tabular}}
\vspace{-0.2cm}
\caption{\footnotesize{\textbf{Offline-to-online fine-tuning on simulated benchmarks}. \methodname{} + Cal-QL outperforms every other approach in aggregate, both in terms of  the offline performance (left of $\rightarrow$) and performance after 1k episodes of fine-tuning (right of $\rightarrow$). This indicates the efficacy of \methodname{} in fine-tuning diffusion policies effectively.}}
\label{tab:offline-online-benchmarks}
\end{table}

\textbf{\emph{Results: \methodname{} with hybrid RL.}} Next, we run \methodname{} on top of RL with Prior Data (RLPD~\citet{ball2023efficient}), a method that incorporates offline data into an online RL training run but does not use offline RL pre-training. In this case, we replace the standard tanh-Gaussian policy used by RLPD with a diffusion policy and keep the critic randomly initialized. As shown in Table~\ref{tab:rlpd-kitchen} (left), observe that \methodname{} is able to improve upon the imitation-learning performance of the diffusion policy after 200 episodes to substantially better performance values than when a Gaussian policy is used for training itself. This further corroborates the efficacy of \methodname{} in efficiently leveraging the expressivity of the policy architecture.

\begin{table}[t]
\centering
\resizebox{0.95\linewidth}{!}{\begin{tabular}{l|cc|cc|cc}
\toprule
\textbf{Task} & \shortstack{ tanh-Gaussian \\ RLPD \\ @ 200} & \shortstack{ Diffusion \\ \methodname{} + RLPD \\ @ 200} &  \shortstack{Gaussian \\ IQL \\ @ 1k}&\shortstack{Diffusion \\. \methodname{} + IQL \\ @ 1k} & \shortstack{tanh-Gaussian \\ Cal-QL \\ @ 1k}&\shortstack{Autoregressive \\ \methodname{} + Cal-QL \\ @ 1k}\\
\midrule
\textbf{partial}& 0 $\to$ 18 & 58 $\to$ 73   & 40 $\to$ 60&62 $\to$ 75 & 59 $\to$ 84&33 $\to$ 95\\
\textbf{mixed}& 0 $\to$ 14 & 58 $\to$ 58 & 48 $\to$ 48&69 $\to$ 73 & 37 $\to$ 72&42 $\to$ 84\\
\textbf{complete}& 0 $\to$ 34 & 70 $\to$ 81  & 57 $\to$ 50&63 $\to$ 88 & 19 $\to$ 57&8 $\to$ 90\\
\bottomrule
\end{tabular}}
\vspace{-0.15cm}
\caption{\footnotesize{\textbf{Combining \methodname{} with different policy classes and critic learning algorithms.} In the hybrid RL setting, \textbf{\methodname{} + RLPD} is able to effectively improve a pre-trained diffusion policy without requiring pre-training the critic. \textbf{\methodname{} + IQL} attains a similar performance on the $\mathrm{FrankaKitchen}$ domain as IDQL, proving our method can work with different objectives for the critic. \textbf{Autoregressive \methodname{}} improves an auto-regressive categorical policy based on a transformer backbone by 224\%. To the best of our knowledge, this is the first time an auto-regressive transformer was improved with the Actor-Critic architecture.}}
\label{tab:rlpd-kitchen}
\vspace{-0.3cm}
\end{table}

\textbf{Results: \methodname{} + Cal-QL with autoregressive categorical policies.} Our next results show that \methodname{} is also effective in training transformer-based policies that model the distribution over actions autoregressively using categorical distributions. Concretely, this type of policy discretizes each dimension of the action space independently into a set of 128 bins, and then trains an autoregressive model over this sequence of discrete per-dimension action tokens. Observe in Table~\ref{tab:rlpd-kitchen} (right) that \methodname{} is also able to effectively improve autoregressive categorical policies with Cal-QL, and attains performance 26\% better than using tanh-Gaussian policies on average across the three tasks considered. This establishes the efficacy of \methodname{} in fine-tuning policies of multiple classes.

\vspace{-0.25cm}
\subsection{Results: RL Fine-Tuning of Robot Policies in the Real World}
\vspace{-0.25cm}
We now show that \methodname{} \emph{can} enable fine-tuning policies on a real robot, resulting in substantial improvements in success rates of the pre-trained policy initialization within just 40 minutes to 2 hours (i.e., 10-70 episodes) of real-world autonomous interaction. To our knowledge, \emph{this is one of the first results to fine-tune diffusion policies and generalist policies on a real robot with value-based actor-critic RL.}

\textbf{Real-world robot and task setup.} We study three manipulation tasks (Figures~\ref{fig:robot_task}, \ref{fig:pot_filmstrip}, and~\ref{fig:teaser}) on a WidowX-250 arm with six degrees of freedom and a single third-person mounted camera. Our setup is inspired by \citet{bridgev1,bridgev2} and the policy controls the end-effector pose at a frequency of 5 Hz for a diffusion policy and 3 Hz for larger autoregressive policies. The tasks that we study are as follows:
\textbf{(a)} \emph{``cup to drying rack''}, which requires grasping a plastic cup and placing it on the drying rack across the sink; \textbf{(b)} \emph{``pot to sink''}, which requires picking and moving a toy pot from the drying rack to the sink; and \textbf{(c)} \emph{``vegetable to sink''}, which requires grasping a toy cabbage and placing it on a plate in the sink. For tasks \textbf{(a)} and \textbf{(c)} the sink contains distractor objects, and for all tasks the position and rotation of the target object are randomized. We collect 10 tele-operated human demonstrations for task \textbf{(a)} and 20 for task \textbf{(b)} to pre-train a diffusion policy and the critic via Cal-QL + \methodname{} that we then fine-tune online. For task \textbf{(b)}, we consider a ``distribution shift'' fine-tuning scenario, where the demonstrations show no distractors, but fine-tuning is supposed to be done with distractor objects. While seemingly benign, this sort of difference between pre-training and fine-tuning setups is still challenging as it leads to poor fine-tuning performance in many prior work that has attempted to run some form of real-robot RL~\citep{kumar2022pre}. For task \textbf{(c)} we leverage a large pre-trained policy to improve performance without any further demonstrations. While OpenVLA~\citep{kim2024openvla} has seen this environment in its dataset, the specific task is new, and there might be small differences in camera angle and background. We collect 50 rollout episodes by zero-shot prompting OpenVLA with the instruction ``put the vegetable on the plate'', and use them to pre-train the critic with Cal-QL + \methodname{}.

\begin{table}[t] %
\vspace{0.1cm}
    \centering
    \resizebox{0.7\linewidth}{!}{\begin{tabular}{l|cc|c}
    \toprule
    \textbf{Task} & \shortstack{DDPM \\ (offline)} & \shortstack{Iterated\\ Filtered BC (online)} & \shortstack{Cal-QL + \methodname{} \\ (offline $\to$ online)}\\
    \midrule
    \textbf{Cup to Rack} & 50\% & 50\% & 55\% $\to$ 90\% \\
    \midrule
    {\textbf{Pot to Sink} (w/ dist. shift)} & 50\% & - & 80\% $\to$ 100\% \\
    \bottomrule
    \end{tabular}}
    \vspace{-0.2cm}
\caption{\footnotesize{\textbf{Real-robot fine-tuning of diffusion policies with \methodname{}.}} \methodname{} improves the performance of an offline pre-trained diffusion policy on two real robot tasks. Notably, while iterated filtered BC, a simple and stable approach for fine-tuning does not meaningfully improve over fine-tuning on task \textbf{(a)}, \methodname{} improves substantially. \methodname{} is similarly effective on task \textbf{(b)}, which fine-tunes and tests with added distractor objects, a common distribution shift in real-world robotics tasks.}
\vspace{-0.3cm}
\end{table}

\begin{figure}[h]
    \vspace{-0.2cm}
    \centering
    \includegraphics[width=0.89\linewidth]{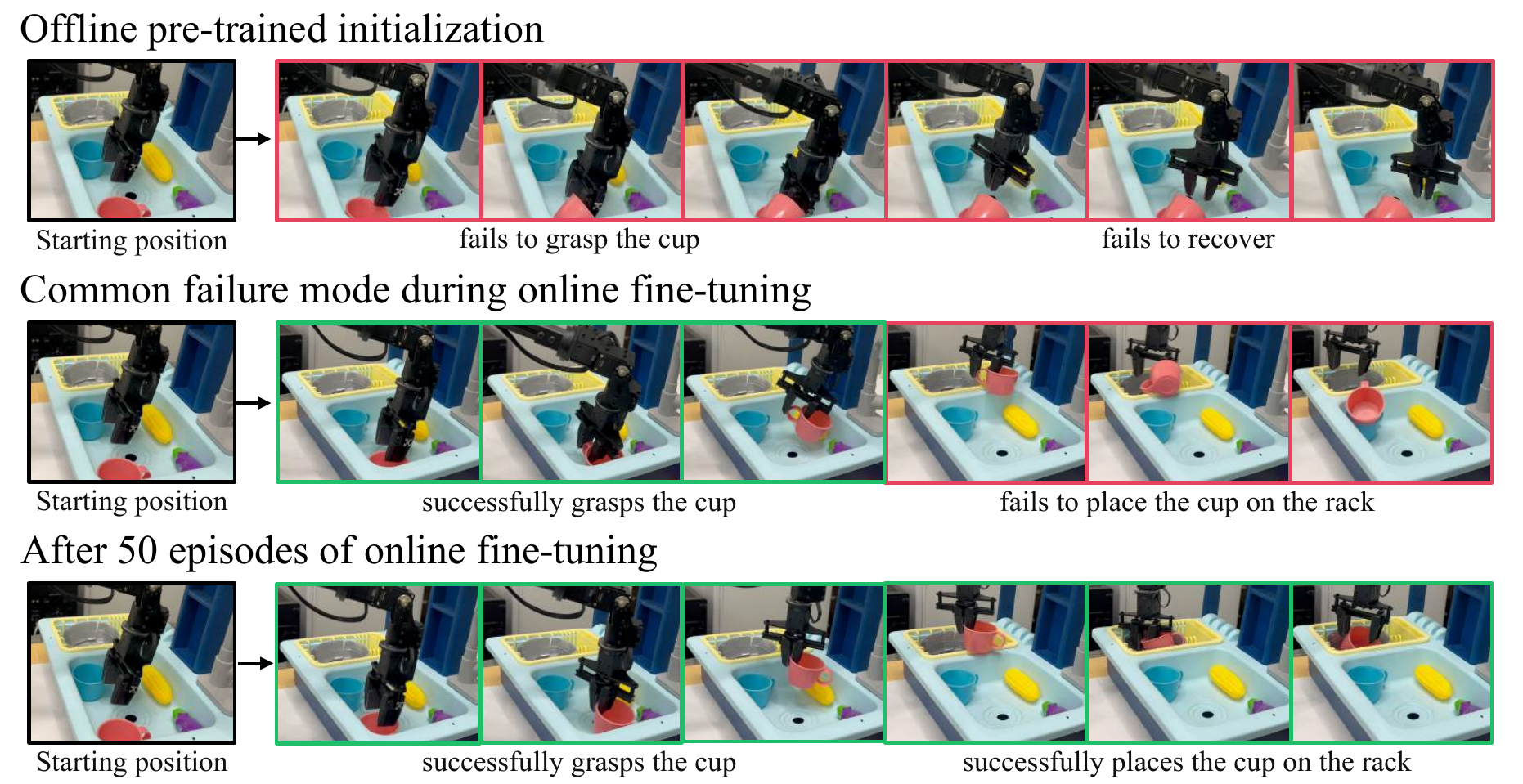}
    \vspace{-0.2cm}
    \caption{\footnotesize{\textbf{Evolution of learned behaviors during online fine-tuning of diffusion policies with \methodname{} on task \textbf{(a)}, with a new initial location for the cup}. The offline initialization (in red) fails to both grasp the cup and place it on the rack. {During intermediate online interaction episodes (in yellow), it successfully grasps the cup, but fails to place it on the rack}. After 50 episodes (in green), it learns to successfully grasp the cup and place it on the rack.}}
    \label{fig:robot_task}
    \vspace{-0.3cm}
\end{figure}
\vspace{-0.25cm}
\subsubsection{Fine-tuning Real-World Diffusion Policies}
\vspace{-0.2cm}
In each case, we fine-tune with a sparse reward function that is based on the detected positions of the target objects and the gripper state. After every robot trial, we perform a manual reset and randomization of the position and orientation of the object.
When running \methodname{} with a diffusion policy on the real robot, we found it important to collect 20 warm-up episodes from the pre-trained offline RL trained policy before updating it. We also compare our approach to a filtered BC for autonomous improvement, based on \citet{zhou2024autonomous} (but without goal conditioning) for one of the tasks \textbf{(task (a))}. We omit this comparison for task \textbf{(b)} since the pre-trained diffusion policy did not produce any success under distribution shift on task \textbf{(b)} for seeding iterative filtered BC. {We also found the diffusion policy to be brittle on task \textbf{(b)}, and we are reporting only the best result it was able to attain.}

\textbf{Real robot diffusion policy fine-tuning results.} We observed significant and efficient performance improvement in both tasks when fine-tuning with \methodname{}, resulting in a 75-100\% higher success rate within 40-110 minutes. We noticed a performance drop during the first 50 episodes of fine-tuning in the \emph{``cup to drying rack''} task, which was consistent with our findings in the $\mathrm{CALVIN}$ task and many other works studying online fine-tuning~\citep{nakamoto2023calql}. We hypothesize that our expressive policy enables the robot to quickly recover and improve within the next 20 episodes. 

\vspace{-0.2cm}
\subsubsection{OpenVLA Fine-Tuning in the Real World} 
\label{sec:openvla}
\vspace{-0.2cm}
Next we fine-tune OpenVLA, a 7B parameter generalist policy. Implementation wise, we had to make some modifications to make it feasible to fine-tune such a large policy autonomously with real-robot RL. First, we discuss some important design decisions for the offline RL stage that trains only a critic. In this phase, we implemented a cache of actions to store actions OpenVLA would take in each state by sampling 16 actions from this generalist policy. This cache enables offline RL critic training in Cal-QL with an OpenVLA policy to still run at similar speeds as a much smaller policy because actions in this cache can be reused for TD backups in the offline RL phase. When coupled with the action optimization phase from \methodname{}, using optimized action particles in the TD backup still allows for policy improvement, even though the OpenVLA policy is not updated in this offline RL phase due to computational cost associated with it. \textbf{During online fine-tuning,} we now update the parameters of the generalist OpenVLA policy. Concretely, we distill optimized actions into OpenVLA via LoRA~\citep{hu2021lowrank} fine-tuning with rank=32 to speed up training.
After policy distillation epochs, we recompute the cache of OpenVLA actions with 12 distributed processes, and use these cached actions for critic training.
Aside from using half the number of samples from the base policy due to memory constraints and reduced learning rate for stability, all hyperparameters are the same as in the experiments above.

\textbf{Results.} After 1 hour of zero-shot trials (where base OpenVLA obtained 40\% success rate) and 40 minutes of online RL fine-tuning, the resulting fine-tuned OpenVLA policy obtained 70\% success rate (Figure~\ref{fig:teaser} (middle)). We observe that the base OpenVLA policy often grasps the wrong object if the gripper is close to the distractor object. After fine-tuning, this error mode is significantly reduced. The fine-tuned policy often grasped the target object more securely, whereas base OpenVLA sometimes let the object fall (please see our project website \href{https://policyagnosticrl.github.io/}{https://PolicyAgnosticRL.github.io/} for evaluation trajectory examples).

\vspace{-0.45cm}
\subsection{Ablation Studies and Controlled Experiments}
\vspace{-0.2cm}

\begin{table}[t]
\centering
\vspace{-0.2cm}
\resizebox{0.6\linewidth}{!}{\begin{tabular}{l|cc|c}
\toprule
Task & \shortstack{\methodname{} \\ no global opt.}& \shortstack{\methodname{} \\ no local opt.}& \methodname{}\\
\midrule
antmaze-large-diverse & 0 $\to$ 0& 74 $\to$ 95&73 $\to$ 93\\
CALVIN & 215 $\to$ 389 & 201 $\to$ 357& 234 $\to$ 455\\
\bottomrule
\end{tabular}}
\vspace{-0.15cm}
\caption{\footnotesize{\textbf{Understanding the importance of global and local optimization.} We compare the performance of \methodname{} + Cal-QL with and without global optimization as measured by average return obtained Note that not using both local and global optimization leads to worse performance. On diverse data such as antmaze-large-diverse, we find global optimization is crucial. On somewhat more narrow data, (e.g., play data in $\mathrm{CALVIN}$) local optimization is also important.}}
\label{tab:ablation-global-search}
\end{table}
Finally, we present some experiments to understand the importance of each component of \methodname{}: \textbf{(1)} when is global optimization (Equation~\ref{eq:global optimization}) important for improving the policy? \textbf{(2)} when is local optimization (Equation~\ref{eq:local_optimization}) important for improving the policy? \textbf{(3)} is using a pre-trained policy for action optimization initialization necessary, or would a strong optimizer (e.g., CEM) suffice from a random initialization?, and \textbf{(4)} is sampling actions from the current policy important for Stage II of \methodname?

\textbf{(1), (2): Effect of global and local optimization.} On the two tasks we study (antmaze-large-diverse and $\mathrm{CALVIN}$), we make a number of interesting observations (see Table~\ref{tab:ablation-global-search}, and Figures~\ref{fig:local-optimization-ablation} and~\ref{fig:global-optimization-ablation} for number of gradient steps and base policy samples ablations). First, we find that both local and global optimization are critical for performance in some environment: on antmaze-large-diverse, global optimization is critical, but local optimization is not as important. On $\mathrm{CALVIN}$, on the other hand, both components are important. This tells us that global optimization is important in general, but local optimization is perhaps only useful when we have a somewhat narrow dataset (e.g., action coverage on $\mathrm{CALVIN}$ is narrow; while action coverage on antmaze is quite high). Thus, \emph{we recommend the general workflow} of always deploying global optimization when running \methodname{} and strongly using local optimization when the dataset action distributions are somewhat narrow.

\begin{figure}[t]
    \centering
    \vspace{-0.3cm}
    \includegraphics[width=0.9\linewidth, trim=0 35 0 0, clip]{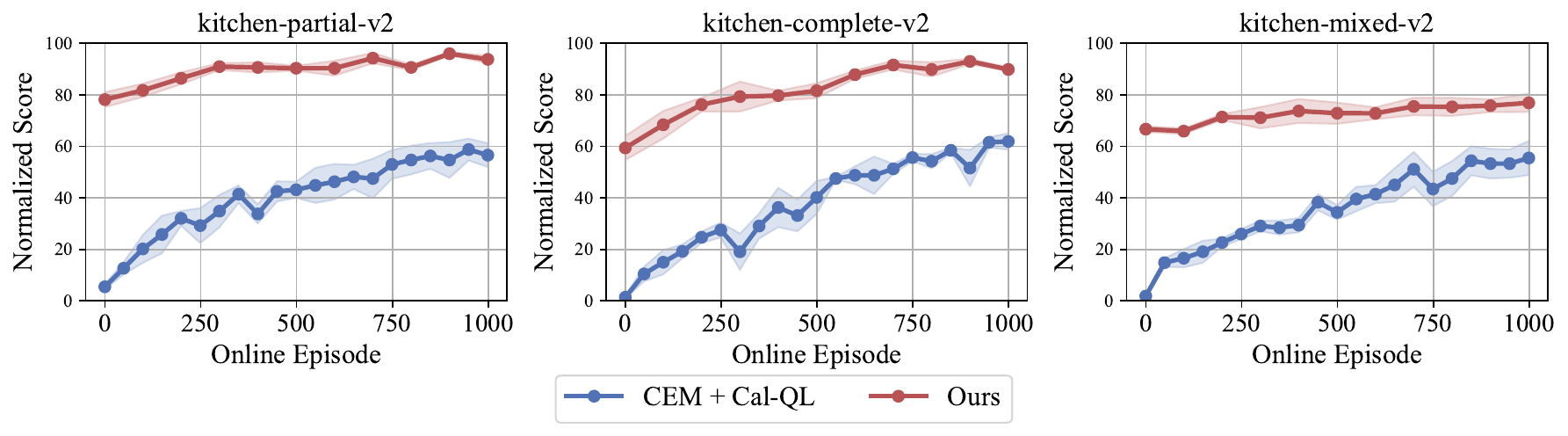}
    \includegraphics[width=0.6\linewidth]{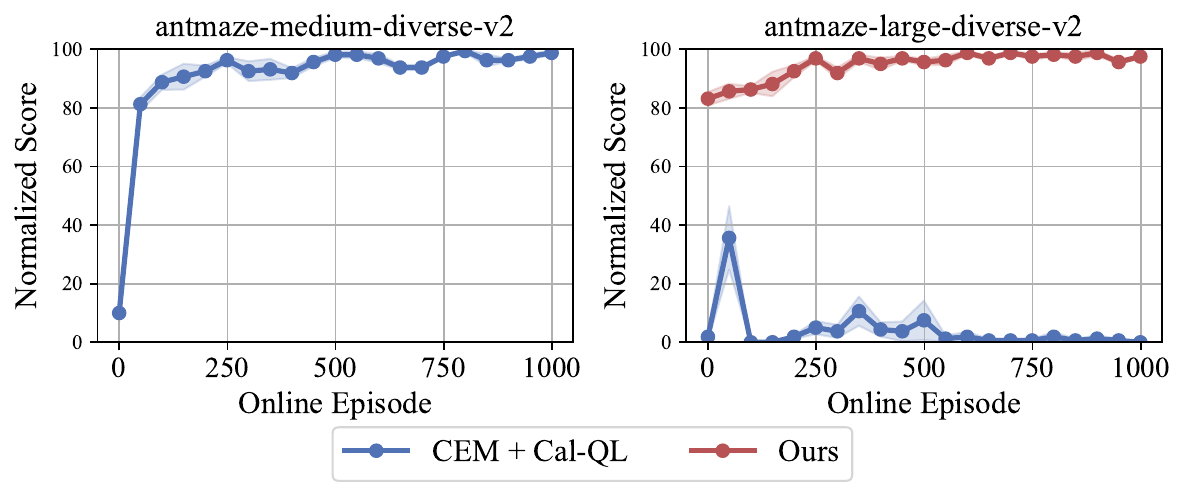}
    \vspace{-0.2cm}
    \caption{\footnotesize{{\textbf{\textit{Comparison with CEM optimizer.}} Instead of using the action optimization procedure detailed in Section~\ref{sec:method}, any time the Cal-QL algorithm queries the policy we perform a Cross-Entropy Method optimization process to obtain actions. We use the same CEM hyper-parameters as~\citet{simmons2019q}, and maintain the Cal-QL hyper-parameters and architectures as \methodname{}. for all tested environments, the performance after pre-training (i.e. at step 0, before taking any online steps) is at or close to 0, and performance improves over the course of fine-tuning, but remaining well below PA-RL with a diffusion policy.}}}
    \label{fig:cem-kitchen-antmaze}
    \vspace{-0.3cm}
\end{figure}

\begin{wrapfigure}{r}{0.4\textwidth}
    \centering
    \vspace{-0.8cm}
    \includegraphics[width=0.99\linewidth]{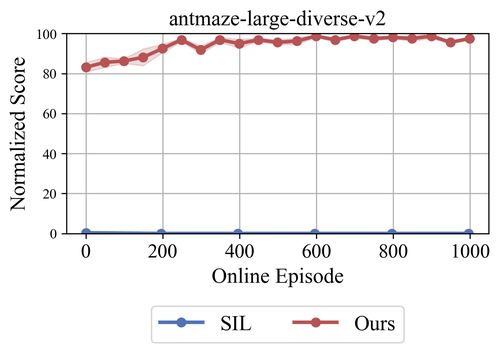}
    \vspace{-0.75cm}
    \caption{\footnotesize{\textbf{Comparison with training on dataset actions} on antmaze-large-diverse-v2. For fairness, critic pre-training and fine-tuning are done in the same manner as \methodname{}.}}
    \label{fig:sil-overestimation}
    \vspace{-0.5cm}
\end{wrapfigure}
\textbf{(3), (4): CEM optimizer and the effect of using the current policy for proposing actions.} The action optimization procedure in \methodname{} seeks to find actions that maximize predicted Q-values within a limited budget for both global and local optimization. This implicitly constrains the action optimization procedure to not deviate substantially far away from the current policy, which is initialized via pre-training on offline data. To understand whether using a pre-trained policy is helpful, or whether simply maximizing Q-values is enough, we replace action optimization with a more powerful optimization procedure, cross-entropy method (CEM), which iteratively refines actions by keeping the top few according to the learned critic, but starts from random actions. Figure~\ref{fig:cem-kitchen-antmaze} shows a comparison in the Antmaze and Kitchen domains. CEM initialized from scratch performs very poorly after pre-training for all tested tasks and reaches significantly lower asymptotic performance. Figure~\ref{fig:cem-overestimation} shows that this is because CEM finds actions where the Q-function greatly overestimates values. This implies that not deviating too far from the data is also important for \methodname{}.

Moving forward, CEM could still leverage a fixed pre-trained policy to limit considered actions to be close to seen actions. To assess the importance of sampling actions from the current snapshot of the learned policy, in Figure~\ref{fig:cem-initialized}, we compare \methodname{} against CEM + Cal-QL, where the CEM optimization procedure is initialized with a fixed diffusion policy pre-trained with imitation learning.
On kitchen-complete-v2 and CALVIN, tasks which exhibit lower coverage of the action space and highly multimodal datasets, \methodname{} still significantly
outperforms CEM. We believe that the data composition on these domains is hurting CEM performance, as CEM can \textit{average} the different modes of behavior. Concretely, a CEM iteration consists of selecting a new action distribution using the mean and variance of the highest ranked under the critic (i.e., CEM assumes a Gaussian distribution). If the pre-trained policy is multimodal, this averaging operation can lose multimodality, and result in a new action ``lying in the middle'' of two modes of the pre-trained policy, but which itself is less likely and out-of-distribution under the pre-trained policy.

\textbf{(4): Effect of using on-policy actions over dataset actions for policy distillation.} \methodname{} is similar to self-imitation learning~\citep{oh2018self} or advantage-weighted regression (AWR)~\citep{peng2019advantage} if we only optimize weighted log likelihoods on one action sample from a stale policy (e.g., the behavior policy of the offline dataset or the replay buffer), turn off local optimization, and the policy distillation loss is weighed by positive action advantages (or exponentiated advantages). We compare against this approach on antmaze-large-diverse-v2 in Figure~\ref{fig:sil-overestimation}. We generically refer to this approach as self-imitation learning (SIL) and note that it fails to get any positive performance on this task. As shown in Figure~\ref{fig:global-optimization-ablation}, taking multiple samples from the base policy is critical for antmaze-large-diverse-v2, which could explain the poor performance of methods that only sample one action from a stale policy for learning, despite using advantage weights.

\vspace{-0.2cm}
\section{Discussion and Conclusion} 
\vspace{-0.2cm}
In this paper, we developed \methodname{}, a method to fine-tune policies of various classes and parameterizations via actor-critic RL. \methodname{} directly optimizes multiple action samples against the critic via re-ranking and gradient ascent to obtain an improved set of actions, that are then used to supervise the policy. We showed state-of-the-art online fine-tuning results across a number of simulation tasks and on two real-robot tasks. Despite promising results, \methodname{} still has some limitations that future work should aim to address. Most importantly, \methodname{} requires sampling multiple actions from the policy, which is expensive for large foundation policies. That said, future work can attempt to reduce this computational cost by caching actions from past rounds and training on them using ideas from off-policy policy gradient. Understanding interplay between global and local optimization better is also a viable direction.
Finally, we also remark that while we utilize a Cal-QL critic for fine-tuning policies, including the generalist OpenVLA policy in Section~\ref{sec:openvla}, this Q-function critic is parameterized by a non-generalist model. An important direction in future work is to develop approaches to train a generalist critic models. 

\vspace{-0.2cm}
\section*{Acknowledgements}
\vspace{-0.2cm}
We thank Zheyuan Hu, Bhavya Agarwalla, Fahim Tajwar, Abitha Thankaraj, Mitsuhiko Nakamoto, Kyle Stachowicz, Guanya Shi, Max Simchowitz, Katerina Fragkiadaki, Abhishek Gupta, and anonymous reviewers for informative discussions and feedback on an earlier version of this work. This work was supported by the Office of Naval Research under grant N00014-24-12206 and OpenAI SuperAlignment Grants, and the Stanford Graduate Fellowship. We thank TPU Research Cloud (TRC), NCSA, and Google Cloud for generous compute donations that made this work possible.

\bibliography{iclr_conference.bbl}

\appendix
\newpage
\part*{Appendices}
\section{Environment details}
\label{appendix:env-details}

\begin{figure}[htbp]
	\centering
	\subfloat[Ant Maze Environment]{
		\includegraphics[width=0.3\textwidth, height=0.2\textheight]{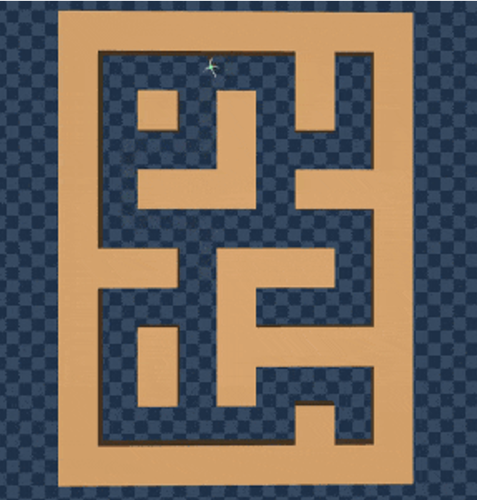}
		\label{fig:ant_maze}
	}
	\hspace{0.01\textwidth}
	\subfloat[Franka Kitchen Environment]{
		\includegraphics[width=0.3\textwidth, height=0.2\textheight]{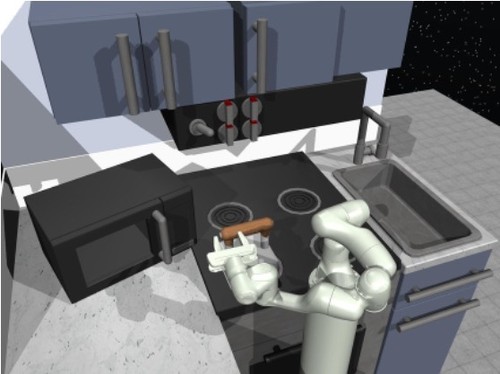}
		\label{fig:franka_kitchen}
	}
	\hspace{0.01\textwidth}
	\subfloat[Calvin Environment]{
		\includegraphics[width=0.3\textwidth, height=0.2\textheight]{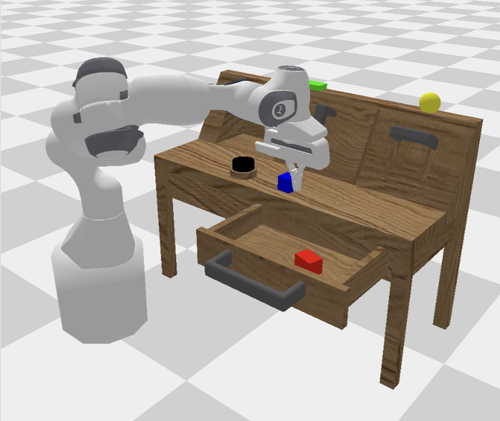}
		\label{fig:calvin}
	}
	\caption{\footnotesize{\emph{\textbf{Simulation Environments}}}}
	\label{fig:environments}
\end{figure}

\textbf{D4RL AntMaze:}
We test methods across two maze sizes ($\mathrm{medium}$ and $\mathrm{large}$) and two dataset types ($\mathrm{play}$ and $\mathrm{diverse}$). The $\mathrm{diverse}$ and $\mathrm{large}$ datasets differ in the starting locations and goal locations of trajectories. The $\mathrm{diverse}$ dataset consists of trajectories with random initial and goal locations, whereas $\mathrm{play}$ contains a set of specific hand-picked locations.  The offline datasets for this benchmark have high coverage over states and actions.

\textbf{D4RL FrankaKitchen:} The $\mathrm{FrankaKitchen}$ benchmark contains three tele-operated datasets: {kitchen-complete}, which contains trajectories that fully solve all sub-tasks, but is 37 times smaller than the other datasets; {kitchen-partial}, where there are both trajectories that fully solve all sub-tasks, and undirected data that performs unrelated behaviors; and {kitchen-mixed}, where no trajectory solves all tasks, requiring exploration from the agent.

\textbf{Calvin:} We use the task setup introduced by~\citet{shi2022skill}, in which the robot arm needs to complete four tasks ($\mathrm{Open Drawer}$, $\mathrm{Turn on Lightbulb}$, $\mathrm{Move Slider Left}$, and $\mathrm{Turn on LED}$), with the distinction that we only use image observations (i.e., the agent does not have access to proprioception nor object states). To ensure Markovian rewards, we make the reward function equal to the number of completed sub-tasks at each time-step (i.e., the agent only gets reward +4 if all sub-tasks are completed). The evaluation score for a trajectory is the maximum number of sub-tasks completed simultaneously at any single point in the trajectory.

Results for all environments and experiments are averaged over 5 random seeds and 32 evaluations per seed at each evaluation time-step (Figure \ref{fig:paper_plots}). Scores are scaled from [0, 4] to [0, 100]. Shaded regions in the plots are standard errors over random seeds.

\section{Experiment Details}

\subsection{Details and hyperparameters for \methodname{}}
\label{appendix:parl-details}

\textbf{Action optimization hyperparameters:} For all experiments shown in the paper except for ablations, the number of actions sampled from the base policy is \textbf{32}, which are filtered down to the top \textbf{ten}, and then propagated through the Q-function for \textbf{ten} gradient steps with gradient step size of \textbf{3e-4}. While we find that these values are robust to all the tested settings, these choices might require changes according to the characteristics of the available dataset and action space. For example, larger action spaces (such as bimanual manipulation) might require larger gradient step sizes or close-to-optimal datasets might perform well with significantly fewer action samples and gradient steps.

\textbf{Distributional critic:} When any of the random seeds in a domain showed instability in the critic pre-training (i.e. had exploding Q-values) we switched the critic from an MLP that predicts the continuous action value to a distributional critic and trained with the HL-Gauss loss~\citep{farebrother2024stop} instead. Specifically, we switched to a distributional critic for the $\mathrm{AntMaze}$ and $\mathrm{FrankaKitchen}$ domains, and we trained with MSE on $\mathrm{Calvin}$ and the real robot experiments.

\textbf{Sampling vs argmax for action candidate selection:} For environments in which CQL/Cal-QL used the max-backup version of Q-target calculation (namely, all 4 $\mathrm{AntMaze}$ environments), we find that taking the argmax of $\pi_\phi^\mathrm{Opt}$ during inference yielded slightly faster convergence than sampling from the considered actions. During \textbf{policy distillation}, to decide whether to imitate only the argmax of $\pi_\phi^\mathrm{Opt}$ or whether to imitate all samples, we keep track of the variance of action candidate Q-values during pre-training. If the variance is too small, we find that training only with the argmax performs better. Otherwise, training with samples from the categorical distribution yields slightly better results.

\begin{table}[h!]
	\centering
	\begin{tabular}{lcc}
		\toprule
		\textbf{Environment} & \textbf{Policy Training Argmax Action} & \textbf{Policy Training Softmax} \\
		\midrule
		kitchen-partial-v2   & 89.375                                 & 95.3125                          \\
		kitchen-complete-v2  & 90.3125                                & 94.53125                         \\
		kitchen-mixed-v2     & 67.96875                               & 75.15625                         \\
		CALVIN               & 60.6771                                & 46.5625                          \\
		\bottomrule
	\end{tabular}
	\caption{\footnotesize{\emph{\textbf{Comparison between doing policy distillation with samples from $\pi_\phi^\mathrm{Opt}$ and only the argmax.}}}}
	\label{tab:policy_distillation_targets}
\end{table}

\begin{table}[h!]
	\centering
	\begin{tabular}{lc}
		\toprule
		\textbf{Environment} & \textbf{STD of Action Candidate Q-values} \\
		\midrule
		kitchen-partial-v2   & 1.56                                      \\
		kitchen-complete-v2  & 2.66                                      \\
		kitchen-mixed-v2     & 11.54                                     \\
		CALVIN               & 0.02                                      \\
		\bottomrule
	\end{tabular}
	\caption{\footnotesize{\emph{\textbf{Standard deviation of the Q-values of action candidates ($\widetilde{\mathcal{A}}^T_{\pi, m}$) during pre-training.}}}}
	\label{tab:std_qvalues}
\end{table}

\textbf{Details for image-based domains:} Following~\citet{yarats2021drqv2} we augment image observations with random shift augmentations of 4 pixels. To mitigate the failure case in which the Q-values for different actions on the same state collapse to the same value, we use the Q-function architecture introduced by~\citet{kumar2022pre}. At every layer of the critic MLP, we concatenate the action vector to the inputs, so that the network places more importance to the actions.

\textbf{Base policy hyperparameters:} We use the same Diffusion Policy architecture and training hyperparameters as IDQL~\citep{hansen2023idql}. In particular, we use batch size 1024, T=5 diffusion steps, cosine beta schedule, the LN\_Resnet architecture with hidden dimension size = 256 and n = 3 blocks. We pre-train the diffusion policy with learning rate decay but with a constant learning rate during fine-tuning.
For image-based domains ($\mathrm{CALVIN}$ and real robot) we use a ResNet 18 encoder trained from scratch.
For the auto-regressive transformer policy, we discretize each action dimension into 128 bins, and do not use discretization for the state observations. We use a transformer architecture with 4 layers, 256 hidden size, 8 heads, and learning rate 3e-5.

\textbf{Reward scale and bias:} To maintain consistency of hyperparameters across all domains, we bias all rewards from the offline dataset and replay buffer such that the maximum possible timestep reward is zero, and other possible rewards are negative. In particular, we use bias = -1 for $\mathrm{AntMaze}$ and real robot, and -4 for $\mathrm{FrankaKitchen}$ and $\mathrm{CALVIN}$.

\textbf{Cal-QL hyperparameters:} We carry over most hyper-parameter choices from Cal-QL: critic architecture and learning rate, discount, mixing ratio. \\

\textbf{Table of hyperparameters:}

\begin{table}[H]
	\centering
	\renewcommand{\arraystretch}{1.2}
	\resizebox{\linewidth}{!}{%
		\begin{tabular}{@{}p{0.5\textwidth}@{\hspace{1em}}c@{}}
			\toprule
			\textbf{Critic LR}                              & 3e-4                                                                                     \\
			\textbf{Discount $\boldsymbol{\gamma}$}         & 0.99                                                                                     \\
			\textbf{Critic batch size}                      & 256                                                                                      \\
			\textbf{Base policy batch size}                 & 1024 (Diffusion Policies), 256 (Transformers)                                            \\
			\textbf{CQL $\alpha$}                           & 0.005 ($\mathrm{AntMaze}$ \& $\mathrm{Kitchen}$), 0.01 ($\mathrm{CALVIN}$ \& Real robot) \\
			\textbf{Mixing ratio}                           & 0.25 ($\mathrm{Kitchen}$), 0.5 (Rest)                                                    \\
			\textbf{Optimizer (critic and base policy)}     & Adam~\citep{kingma2014adam}                                                              \\
			\textbf{Critic pre-training grad steps}         & 1e6 ($\mathrm{AntMaze}$), Rest: 5e5                                                      \\
			\textbf{Base policy grad steps}                 & \shortstack{Diffusion policies: 3e6                                                      \\ Transformers: 2e6} \\
			\textbf{Base policy distillation learning rate} & 1e-5 ($\mathrm{kitchen}$), 5e-5 (Rest)                                                   \\
			\textbf{Critic hidden layer sizes}              & [256, 256, 256, 256] ($\mathrm{AntMaze}$), [512, 512, 512] (Rest)                        \\
			\bottomrule
		\end{tabular}}
\end{table}

\subsection{Details and hyperparameters for baselines}

\textbf{IDQL}
We use the IDQL-Imp version of IDQL, in which the Q-function, the value function, and the diffusion policy are fine-tuned with new experiences. We use the same network architectures as \methodname{}. For the IQL $\tau$ expectile, we use 0.9 for $\mathrm{AntMaze}$ and 0.7 for everything else. We remark that results for IDQL are not entirely comparable to their paper because \citet{hansen2023idql} used the ``-v0'' antmaze datasets from D4RL, but \citet{fu2020d4rl} deprecated the ``-v0'' datasets in favor of ``-v2'' due to a bug associated with termination flags in -v0 datasets.

\textbf{DQL}
We extensively tuned DQL for fine-tuning in the absence of any official fine-tuning results. For the main $\eta$ RL weight hyperparameter, we performed an environment-specific hyperparameter search at the pre-training phase, selected the one that performed best, and then kept $\eta$ fixed for fine-tuning. For $\mathrm{AntMaze}$ tasks we tried $\eta = \{0.05, 0.5, 1, 3, 3.5, 5, 7, 9, 11, 13, 15\}$. We chose $\eta=11$ for large-diverse, $\eta=15$ for large-play, $\eta=9$ for medium-diverse, and $\eta=7$ for medium-play.
For $\mathrm{FrankaKitchen}$ tasks we tried $\eta = \{0.005, 0.01, 0.05, 0.1\}$. For partial, complete, and mixed, we chose $\eta=0.005$. For $\mathrm{CALVIN}$ we tried $\eta=\{0.01, 0.1, 1, 5, 10, 15\}$. We picked $\eta=0.01$. For offline checkpoint selection, we follow the original methodology of selecting the checkpoint with second lowest DDPM loss, saving checkpoints every 50k gradient steps.\\

\textbf{Cal-QL}
Since we branch off our hyperparameter choices from Cal-QL, this baseline shares most of \methodname{}'s hyperparameters. We used (256, 256) hidden sizes for the policy architecture for every environment.

\textbf{DPPO}
We train a diffusion-based PPO policy based on a DPPM model pretrained on an offline dataset in each simulated task. For the state-based tasks $\mathrm{AntMaze}$ and $\mathrm{FrankaKitchen}$, we train DPPO-MLP with 40 parallelized environments and an action chunking size of 6 for $\mathrm{AntMaze}$ and 8 for $\mathrm{FrankaKitchen}$. For the pixel-based task $\mathrm{CALVIN}$, we train DPPO-ViT-MLP with 50 parallelized environments and an action chunking size of 4.

\textbf{RLPD}
For Table~\ref{tab:rlpd-kitchen}, we train a gaussian policy from scratch with UTD ratio of 10 (same as with Diffusion \methodname{} + RLPD), critic ensemble size ten, and critic ensemble subsample size of two.

	\section{Filmstrips for Real-World Fine-Tuning of OpenVLA with \methodname{}}
	\label{appendix:openvla}

	\begin{figure}[H]
		\centering
		\begin{subfigure}[t]{0.51\textwidth}
			\centering
			\includegraphics[width=\textwidth]{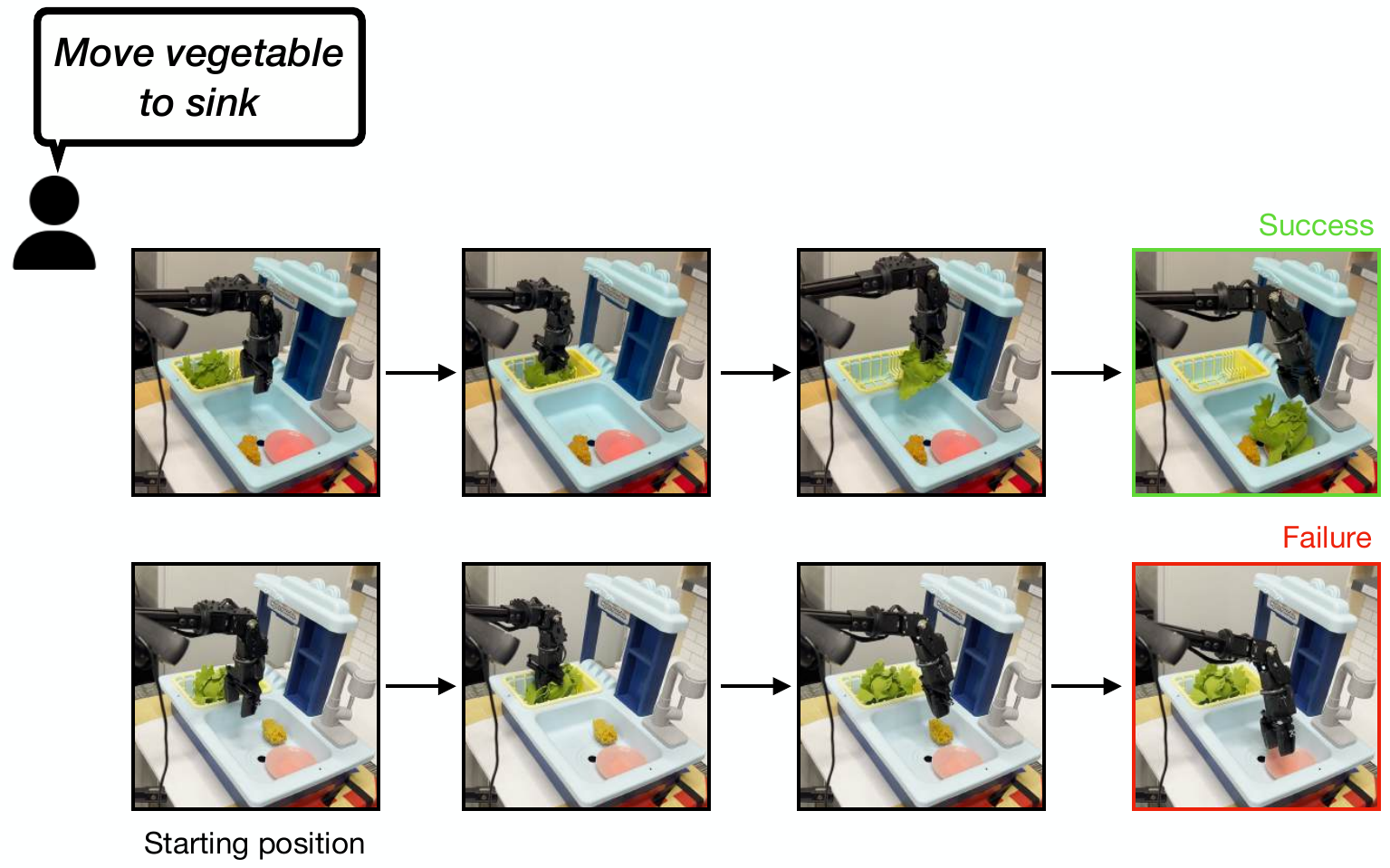}
			\caption{\footnotesize{Zero-shot language-based trials with OpenVLA}}
			\label{fig:openvla-task}
		\end{subfigure}
		\hspace{0.5cm}
		\begin{subfigure}[t]{0.39\textwidth}
			\centering
			\includegraphics[width=\textwidth]{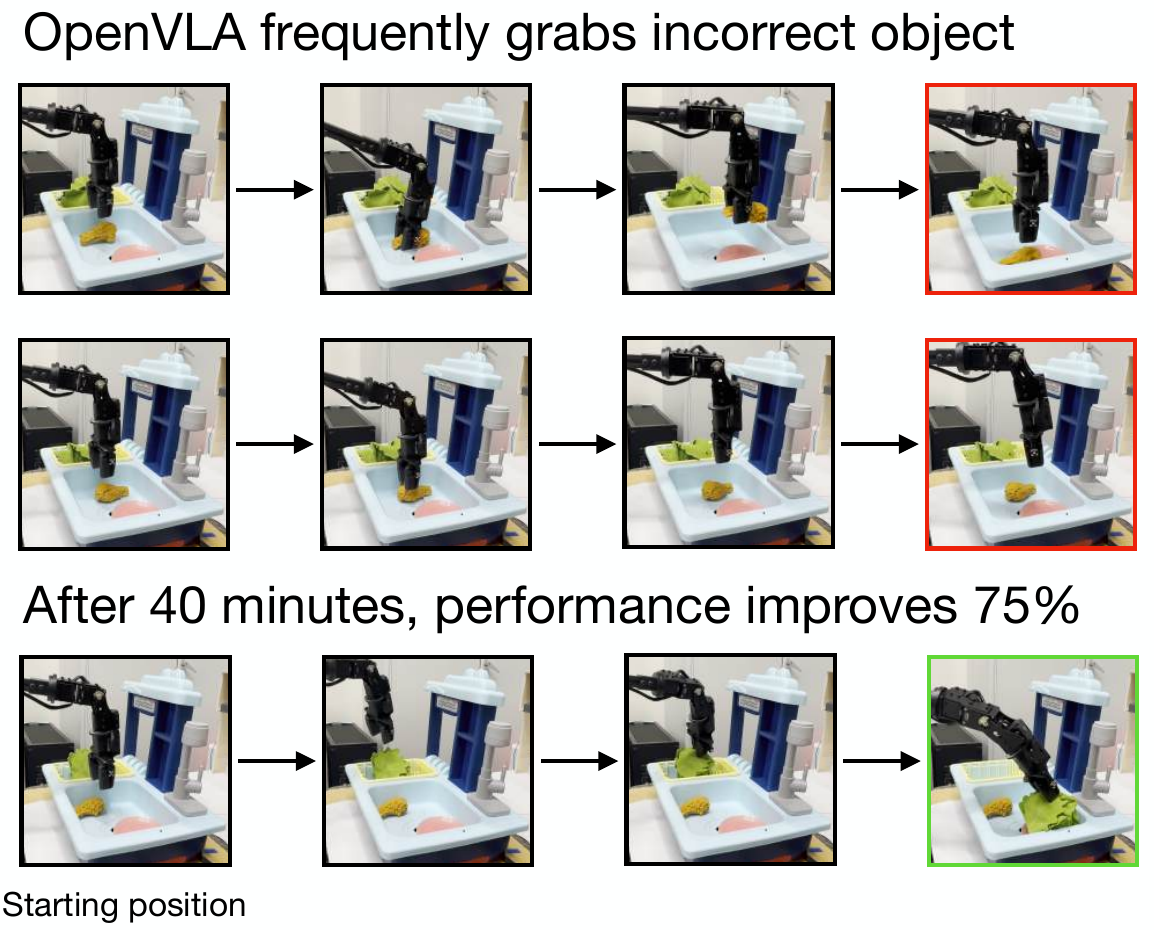}
			\caption{\footnotesize{Online fine-tuning with \methodname{}}}
			\label{fig:openvla-finetuning}
		\end{subfigure}
		\caption{\footnotesize{\textbf{Filmstrips of the manipulation task we fine-tune OpenVLA on.} (Left) the new task, ``vegetable to sink'', requires identifying the vegetable from the distractor (a fried chicken wing), grasping it, and placing it on the pink plate. We collect 50 trials by zero-shot prompting OpenVLA to solve the task. 40\% of the trials are successful. (Right) we deploy \methodname{} to improve OpenVLA for this task, interacting on the real-robot. We observe that OpenVLA frequently grasps the distractor object instead of the vegetable. After 40 minutes of wall clock time, we evaluate the resulting fine-tuned policy. OpenVLA + \methodname{} attained a 70\% success rate.}}
		\label{fig:openvla-figure}
		\vspace{-0.3cm}
	\end{figure}

\section{Additional Figures}
\label{appendix:additional-figures}

\subsection{Real Robot Fine-tuning on task \textbf{(b)}}

\begin{figure}[H]
	\centering
	\includegraphics[width=\textwidth]{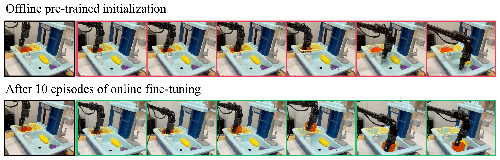}
	\caption{\footnotesize{\textbf{Evolution of learned behaviors during autonomous online finetuning of \methodname{} on task \textbf{(b)} on a difficult pot placement}. The offline initialization (in red) fails to grasp the pot, and gets stuck when attempting to move it to the sink. After only 10 online fine-tuning episodes (in green), \methodname{} learns to successfully complete the task.}}
	\label{fig:pot_filmstrip}
\end{figure}

\subsection{Learning Curves for Auto-Regressive Transformers and IQL with \methodname{}}

\begin{figure}[H]
	\label{fig:transformer_iql}
	\centering
	\includegraphics[width=\linewidth]{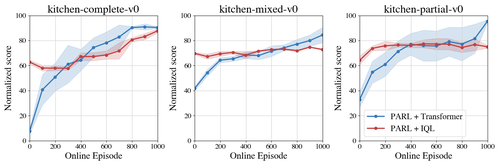}
	\caption{\footnotesize{\textbf{Learning Curves for auto-regressive transformer based policies with \methodname{} and Cal-QL, and diffusion policies with \methodname{} and IQL.}}}
	\label{fig:transformer-iql}
\end{figure}

\subsection{Local and Global Optimization Ablation Experiments}

\begin{figure}[H]
	\centering
	\includegraphics[width=0.9\linewidth]{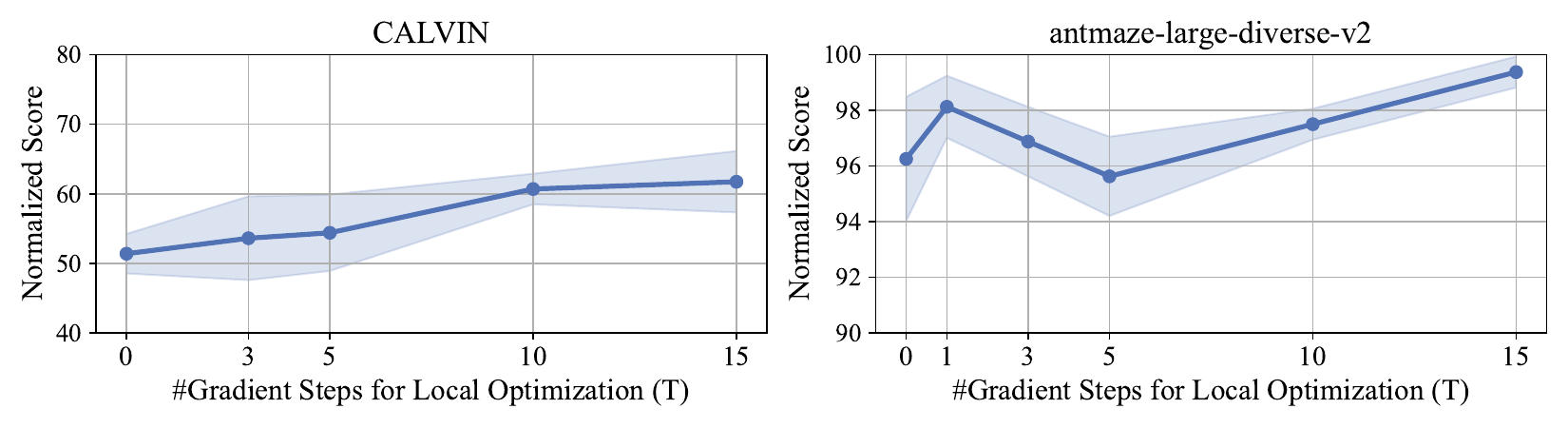}
	\caption{\footnotesize{\textbf{Ablation for the number of gradient steps for local optimization (T).} We plot the evaluation performance for \methodname{} + Diffusion Policy at the end of a fine-tuning budget of 1k episodes on CALVIN (left) and antmaze-large-diverse-v2 (right), taking different numbers of gradient steps during the Local Optimization procedure. We chose to analyze the effect of local optimization on these two tasks because they sit on opposite sides of the data coverage spectrum: CALVIN features relatively little coverage over actions, since the provided dataset is "play data", while antmaze-large-diverse-v2 provides high-coverage over actions (as measured by delta x, delta y, which is more relevant to the task). (Left) CALVIN benefits significantly from increased number of gradient steps, getting up to 20\% increase in final performance compared to taking no gradient steps. (Right) antmaze-large-diverse-v2 already reaches 96\% success rate without taking any gradient steps (i.e., without the local optimization step). We hypothesize that because of the high-coverage, using global optimization with a large-enough number of samples from the base policy already recovers good actions.}}
	\label{fig:local-optimization-ablation}
\end{figure}

\begin{figure}[H]
	\centering
	\includegraphics[width=0.9\linewidth]{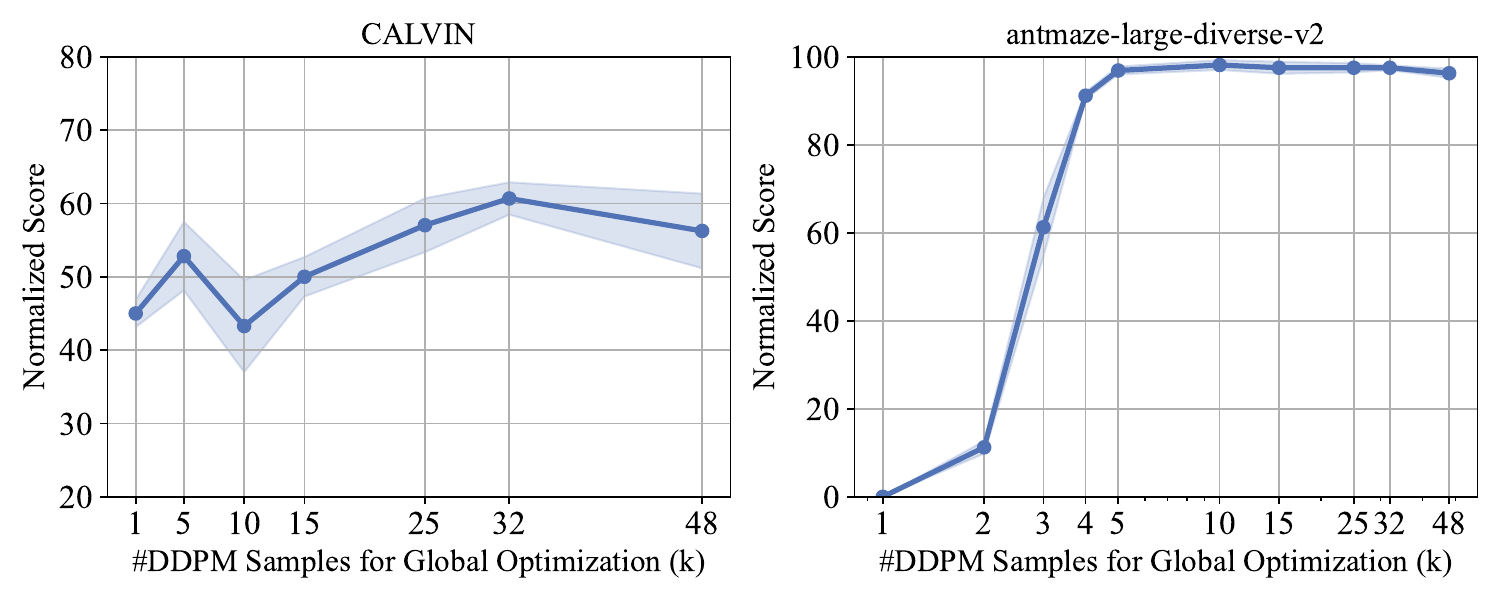}
	\caption{\footnotesize{\textbf{Ablation for the number of samples from the base policy (k).} We plot the evaluation performance for \methodname{} + Diffusion Policy at the end of a fine-tuning budget of 1k episodes on CALVIN (left) and antmaze-large-diverse-v2 (right), sampling different number of actions from the base policy to generate action candidates both for policy distillation and during inference. (Left) CALVIN benefits significantly from increased number of samples from the base policy, attaining 33\% higher normalized score when taking 32 samples (the default value used for \methodname{}) from the policy compared to only 1 sample. (Right) antmaze-large-diverse-v2 exhibits a sharp decrease in final performance when taking fewer than 5 samples from the base policy.}}
	\label{fig:global-optimization-ablation}
\end{figure}

\begin{figure}[H]
	\centering
	\includegraphics[width=0.9\linewidth]{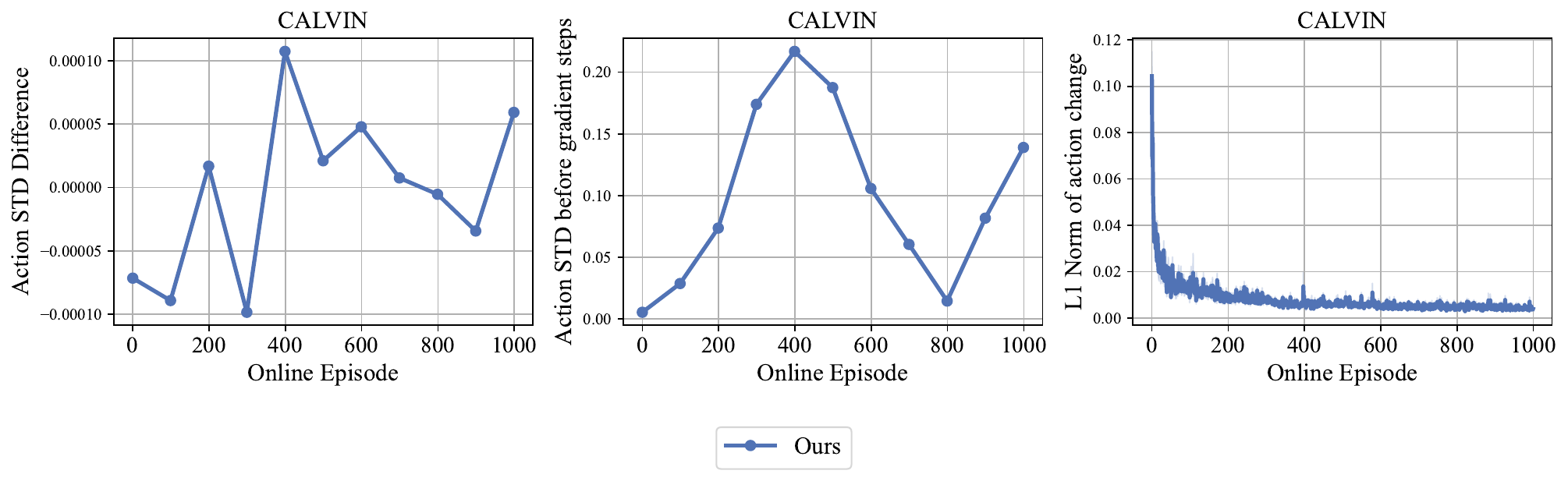}
	\caption{\footnotesize{\textbf{Analysis of the effects of local optimization.} To test whether local optimization results in duplicated action samples, we plot the difference between the standard deviation of action samples before and after taking gradient steps (left) during evaluation episodes on the CALVIN task throughout fine-tuning. The difference in standard deviations is extremely low throughout training. Further, to ensure action samples were not largely duplicates to begin with, and to put the value scale into perspective, we plot the raw standard deviation of action samples before taking gradient steps (center). Standard deviation of actions changes by less than 0.1\% on average during training. Thus, local optimization does not lead to action sample duplication. (Right) we plot the L1-Norm of the change in actions by the local optimization procedure (i.e. the L1 norm of the difference in actions before and after the gradient steps). The biggest direct effect on actions happens in the beginning of fine-tuning, and it quickly decays throughout online training. Note that because of policy distillation, action changes from the local optimization step are compounding (i.e., the actions before applying the gradient steps have already been optimized in past iterations). This might explain the decay in action changes from local optimization.}}
	\label{fig:local-optimization-action-std}
\end{figure}

\subsection{CEM Optimizer + Random Initialization Comparisons}

\begin{figure}[H]
	\centering
	\includegraphics[width=0.9\linewidth]{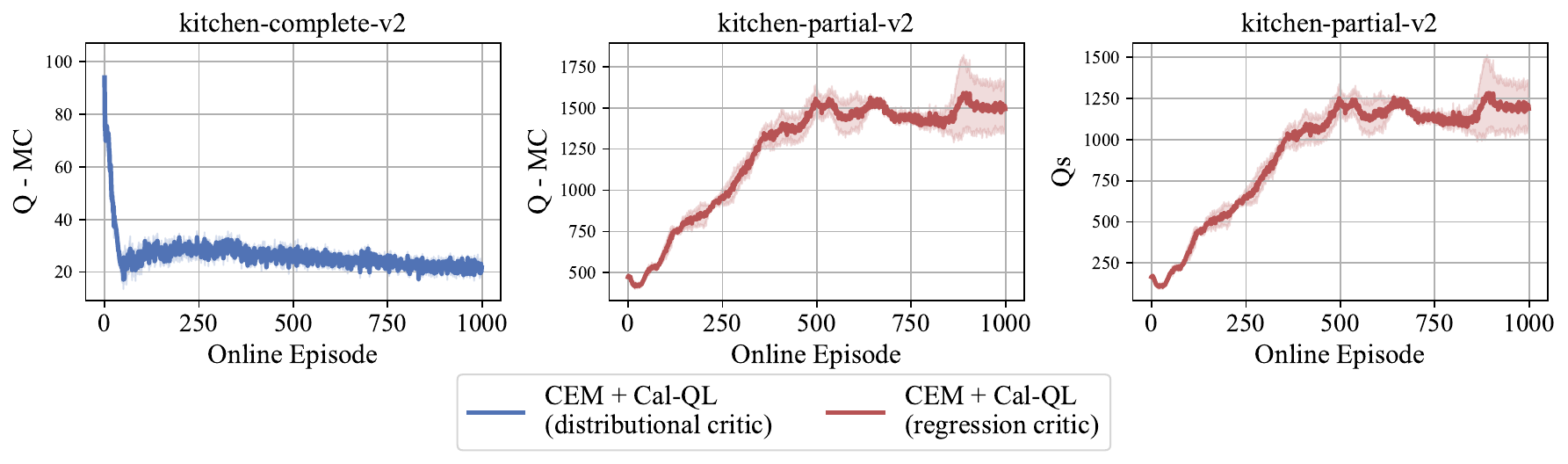}
    \vspace{-0.4cm}
	\caption{\footnotesize{\textbf{CEM exploits Q-function over-optimism.} (Left) We plot the difference between predicted Q-values of CEM actions, and the Monte-Carlo discounted returns that those actions actually got, on kitchen-complete-v2, a task whose dataset contains optimal actions. The critic is trained in the same manner as in Figure~\ref{fig:cem-kitchen-antmaze}. We observe that at the beginning of fine-tuning, predicted Q-values are much higher than the MC returns, even much higher than the predicted Q-values further into training, when task performance is much higher (see Figure~\ref{fig:cem-kitchen-antmaze}). This points to the fact that the CEM optimizer is able to find actions that maximize the Q-function, but are not actually good. (Center) We repeat the same experiment but with a regression-trained critic instead of a distributional critic trained with HL-Gauss. The distributional critic bounds the predicted values by design, which limits over-estimation. By training a Cal-QL critic without a fixed value range (on kitchen-partial-v2), we see much larger over-estimation of Q-values. Predicted Q-values become large positive numbers (right), where rewards are always non-positive.}}
	\label{fig:cem-overestimation}
\end{figure}

\subsection{CEM Optimizer + pre-trained policy initialization}

\begin{figure}[H]
	\centering
	\includegraphics[width=0.8\linewidth, trim=0 35 0 0, clip]{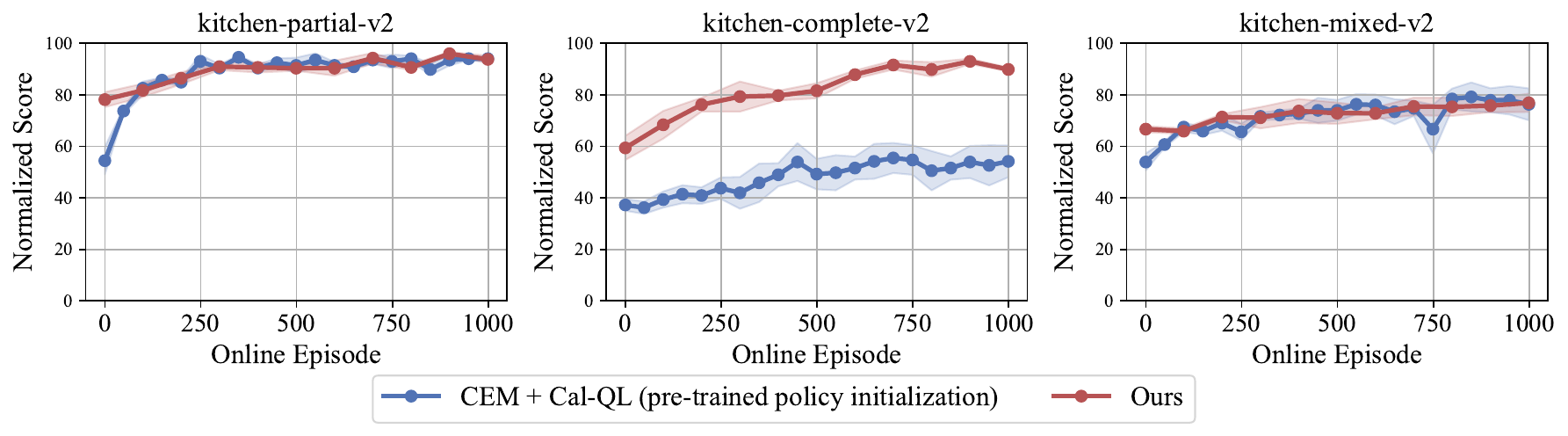}
	\includegraphics[width=0.5\linewidth]{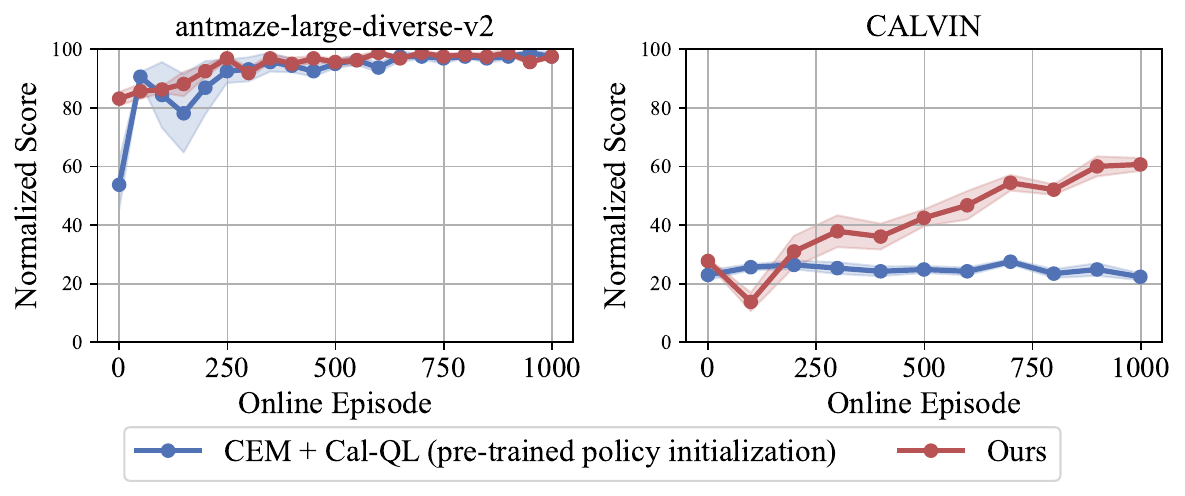}
	\caption{\footnotesize{\textbf{Comparison with CEM optimizer with a pre-trained policy initialization.} We compare to using a CEM optimization procedure where the initial population of actions comes from the same pre-trained policy used for \methodname{}. \methodname{} results in 42\% better offline-only performance across tested domains. In \textrm{antmaze-large-diverse-v2}, \textrm{kitchen-partial-v2}, and \textrm{kitchen-mixed-v2}, CEM quickly catches up and ends with very similar asymptotic performance. In \textrm{kitchen-mixed-v2} and \textrm{CALVIN} \methodname{} significantly outperforms CEM, with 66\% and 172\% better performance respectively. \textrm{kitchen-complete-v2} and \textrm{CALVIN} have lower coverage of actions in their datasets, and \textrm{CALVIN} has highly multi-modal data. We hypothesize these dataset characteristics, which are highly common in real-world robotics datasets, are hurting CEM performance, since CEM can average the different modes of behavior, resulting in OOD actions. Further, CEM lacks an equivalent of the local optimization step to direct exploration towards actions the critic rates highly.}}
	\label{fig:cem-initialized}
\end{figure}

\subsection{Comparison with computing actions for Bellman Backup with the base policy}
\begin{figure}[H]
	\centering
	\includegraphics[width=0.5\linewidth]{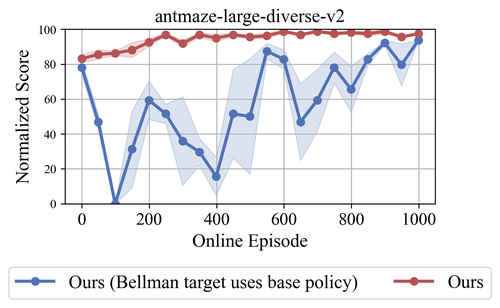}
	\caption{\footnotesize{\textbf{Ablation for the choice of using the optimized action for Bellman backups.} To ablate the choice of computing targets using the optimized policy $\pi_{(\phi, \theta)}^\mathrm{Opt}(\cdot|\cdot, m)$, we compare it against directly sampling from the base policy $\pi_\phi$, and test it on antmaze-large-diverse-v2 fine-tuning. Both methods start from the same pre-trained critic checkpoints. Using the base policy for Bellman targets makes fine-tuning much more unstable, with a sharp drop in performance in the beginning, but ultimately obtains similar performance.}}
	\label{fig:bellman-actions-base-policy}
\end{figure}

\subsection{Learning curves for Gaussian Policies with PA-RL}
\begin{figure}[H]
	\centering
	\includegraphics[width=0.9\linewidth]{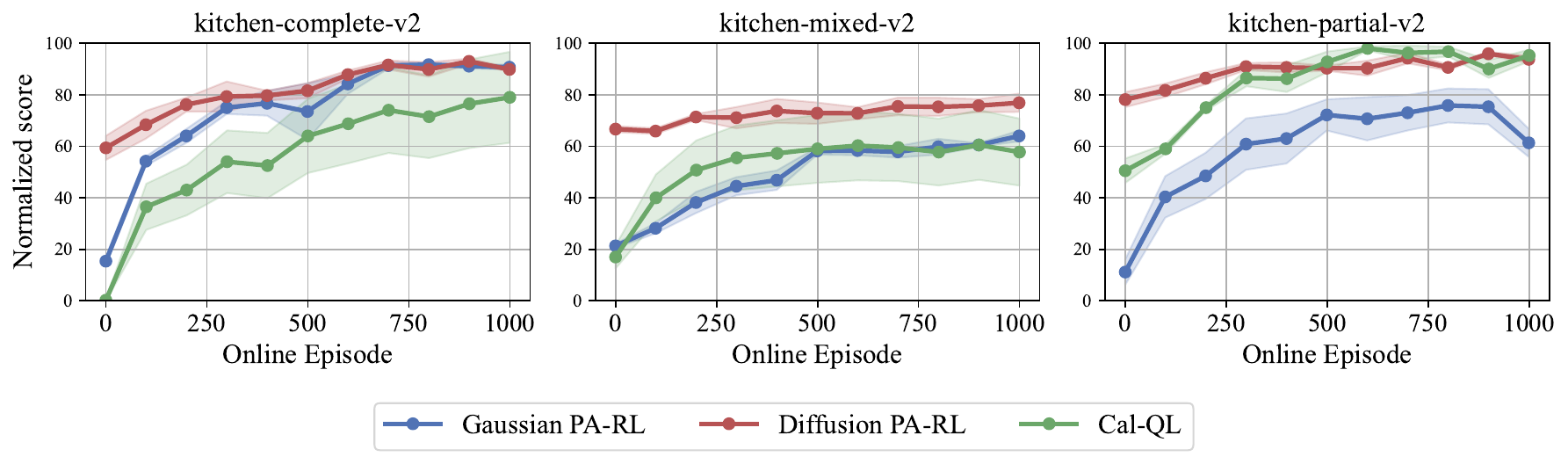}
	\caption{\footnotesize{\textbf{Learning curves for gaussian policies with \methodname{}, compared with Diffusion Policies with \methodname{} and the standard Cal-QL with gaussian policies.} As with other experiments, we first train the base gaussian policy with BC on each dataset, and then do critic pre-training, followed by online RL fine-tuning. The only hyper-parameter we change for gaussian policies is the distillation learning rate, setting it to 3e-4. We observe Gaussian \methodname{} performs competitively with the standard Cal-QL on kitchen tasks.}}
	\label{fig:gaussian-kitchen}
\end{figure}

\section{Training time discussion}
\methodname{} optimizes actions using the procedure described in Section~\ref{sec:method} any time an action from the policy is needed. We discuss how this affects the App of our method at different stages.

\begin{figure}[h]
	\centering
	\includegraphics[width=0.4\linewidth]{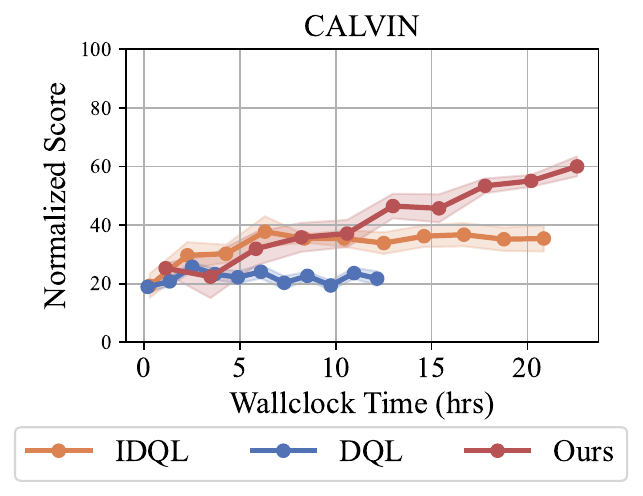}
	\caption{\footnotesize{\textbf{Performance on CALVIN task as a function of wall clock time for \methodname{}, IDQL, and DQL.} All three methods ran on the same compute instence type (TPU v4), were implemented in the same codebase. Observe that \methodname{} improves at a similar rate per unit amount of wall-clock time as IDQL, but is able to improve far beyond to a better performance value. DQL largely remains flat as a function of more unit wall-clock time put into training.}}
	\label{fig:calvin-wallclock}
\end{figure}

\textbf{Critic training.} In principle, action optimization should increase memory and computation requirements to critic training, but it also enables using an action cache to compute ahead of time, even in a distributed manner, when sufficient numbers of actions from the base policy are available. To make sure that this cache is not stale and to ensure that the critic models the optimal / on-policy value function, the actions cache is updated after every epoch of policy training via supervised learning. When sampling from the base policy is more than T times more expensive than taking T gradient steps of the critic (as is the case with OpenVLA or with diffusion policies with a large number of denoising steps), \methodname{} can be significantly more efficient than alternatives that do not do caching.

\textbf{Policy distillation.} Compared to standard offline RL and online fine-tuning objectives, the supervised learning objective \methodname{} can be significantly more efficient than policy improvement through reparameterization. For example, for a diffusion policy, backpropagating critic gradients through the diffusion chain uses a larger memory footprint than the DDPM objective \methodname{} uses, by a factor equal to the number of denoising steps.

\textbf{Inference.} During inference, \methodname{} can optionally also apply action optimization by querying the base policy multiple times to sample an action. This can significantly increase the memory requirements of our method. That said, we do note that the number of samples from the base policy during inference can be much smaller than during training, as we do with OpenVLA (see Appendix~\ref{appendix:openvla}). \methodname{} additionally requires taking multiple gradient steps of the critic with respect to the actions. We note that depending on the architecture used, this can be much cheaper than doing multiple full forward passes through the Q-function. For example, for image-based domains, the bulk of the computation happens for image encoding, which does not depend on the action. Therefore, the gradient steps will ignore that part of the network. There is also room for improvement for future work to investigate reducing the number of gradient steps further into training (as Figure~\ref{fig:local-optimization-action-std} right suggests local optimization might have diminishing effects as fine-tuning progresses).

\end{document}